\documentclass[10pt,twocolumn,letterpaper]{article}

\usepackage[pagenumbers]{cvpr} %

\pdfoutput=1
\usepackage[dvipsnames, table]{xcolor}
\usepackage{color}
\usepackage{tikz,pgfplots,pgfplotstable}
\usepackage{float}
\usepackage{amsthm}
\usepackage{diagbox}
\usepackage{xspace}
\usepackage{multirow}
\usepackage{tabularx}
\usepackage{xr}
\usepackage{algorithm} 
\usepackage{algorithmic}  
\usepackage[algo2e]{algorithm2e} 
\usepackage{pifont}
\usepackage{minitoc}        %
\usepackage{comment}
\usepackage{pstricks}

\newcommand{\metric}[0]{worst category-wise testing error\xspace}

\newcommand{\cmark}{\ding{51}}%
\newcommand{\xmark}{\ding{55}}%
\definecolor{ClrHighlight}{RGB}{210, 255, 255}
\definecolor{green01270}{RGB}{80, 200, 120}
\definecolor{ForestGreen}{RGB}{34,139,34}
\definecolor{darkgray}{RGB}{113, 121, 126}

\newtheorem{definition}{Definition}

\newcommand{\Sec}[1]{{Sec.}~#1}
\newcommand{\Tab}[1]{{Tab.}~#1}
\newcommand{\Fig}[1]{{Fig.}~#1}

\newcommand{\Eq}[1]{{Eq.}~#1}
\newcommand{\Alg}[1]{{Alg.}~#1}

\newcommand{\Lss}[1]{$\mathcal{L}_{\mathrm{#1}}$}
\newcommand{\inc}[1]{(\textcolor{red}{$\uparrow#1\%$})}

\usetikzlibrary{fit,calc}

\colorlet{codehighlightblue}{cyan!60}

\definecolor{cvprblue}{rgb}{0.21,0.49,0.74}
\definecolor{strongercvprblue}{rgb}{0,0.20,0.74}
\usepackage[pagebackref,breaklinks,colorlinks,citecolor=cvprblue,linkcolor=strongercvprblue]{hyperref}

\usepackage{pifont}

\def\paperID{0} %
\def\confName{0}
\def\confYear{0}
\def\paperTitle{R.I.P.~\includegraphics[height=15pt]{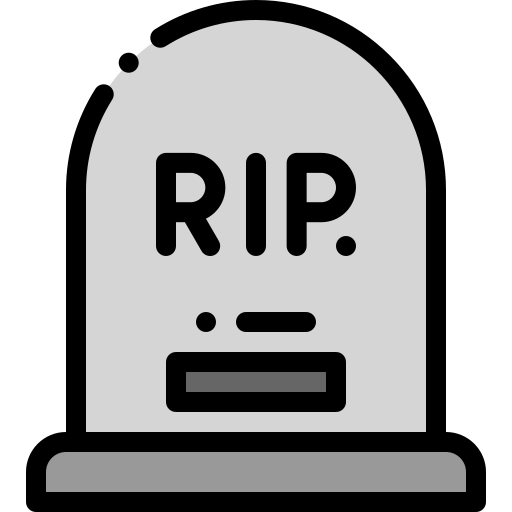}: A Simple Black-box Attack on Continual Test-time Adaptation}
\title{\paperTitle}

\author{Trung-Hieu Hoang$^1$ \qquad Duc Minh Vo$^2$ \qquad Minh N. Do$^1$\\
$^1$Department of Electrical and Computer Engineering, University of Illinois at Urbana-Champaign, USA\\
$^2$The University of Tokyo, Japan\\
{\tt\small \{hthieu, minhdo\}@illinois.edu, \qquad
\tt vmduc@nlab.ci.i.u-tokyo.ac.jp
}
}

\begin{document}
\maketitle

\doparttoc %
\faketableofcontents %

\begin{abstract}
    Test-time adaptation (TTA) has emerged as a promising solution to tackle the continual domain shift in machine learning by allowing model parameters to change at test time, via self-supervised learning on unlabeled testing data. At the same time, it unfortunately opens the door to unforeseen vulnerabilities for degradation over time. Through a simple theoretical continual TTA model, we successfully identify a risk in the sampling process of testing data that could easily degrade the performance of a continual TTA model. We name this risk as \textbf{\underline{R}eusing of \underline{I}ncorrect \underline{P}rediction (RIP)} that TTA attackers can employ or as a result of the unintended query from general TTA users. The risk posed by RIP is also highly realistic, as it does not require prior knowledge of model parameters or modification of testing samples. This simple requirement makes RIP as the first \textbf{black-box} TTA attack algorithm that stands out from existing white-box attempts. We extensively benchmark the performance of the most recent continual TTA approaches when facing the RIP attack, providing insights on its success, and laying out potential roadmaps that could enhance the resilience of future continual TTA systems.

\end{abstract}
\setlength{\abovedisplayskip}{4.5pt}
\setlength{\belowdisplayskip}{4.5pt}
\setlength{\abovedisplayshortskip}{4.0pt}
\setlength{\belowdisplayshortskip}{4.0pt}

\section{Introduction}
\label{sec:introduction}
\begin{figure}[t!]
    \centering
    \includegraphics[width=.9\linewidth]{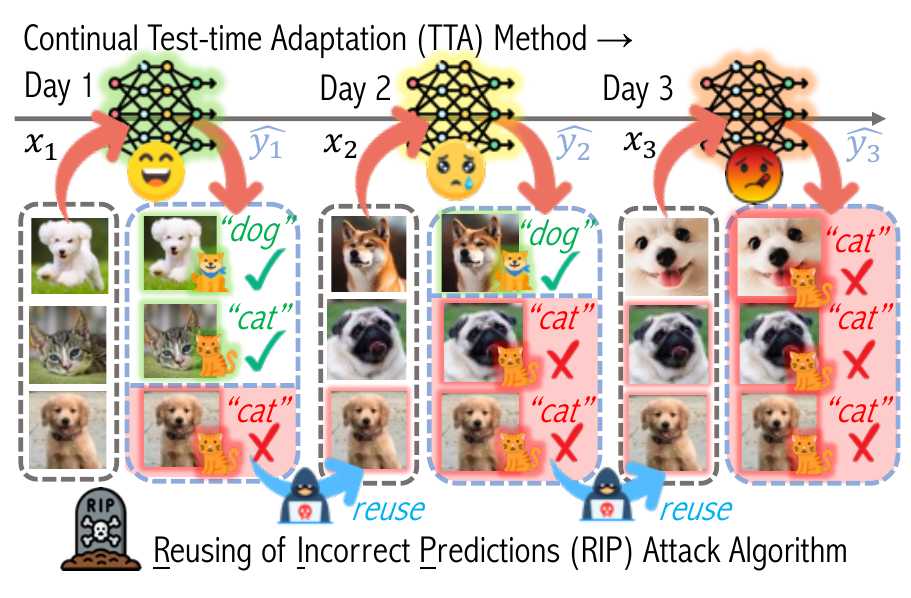}
    \vspace*{-1.0\baselineskip}
    \caption{An illustration of our \underline{R}eusing of \underline{I}ncorrect \underline{P}redictions (RIP) attack against a continual test-time adaptation (TTA) method. Here the attacker intentionally reuses samples that were incorrectly predicted in the subsequent rounds to make the model more confident in these erroneous predictions. 
    RIP is the first black-box attack that can realistically collapse a TTA model. }
    \label{fig:overview}
    \vspace*{-1.5\baselineskip}
\end{figure}

\textit{``No man ever steps in the same river twice, for it's not the same river and he's not the same man".}
\begin{flushright}%
-- \textit{Heraclitus (535 B.C. to 475 B.C.)} -- 
\end{flushright}%
This age-old statement holds in modern machine learning (ML) research, where the real-world environment, like a \textit{river}, is constantly changing, necessitating adaptation to continual domain shifts~\cite{10.5555/1462129, pmlr-v37-ganin15, Hoang_2024_CVPR}.
Recently recognized continual\textit{ Test-Time Adaptation (TTA)}~\cite{wang2021tent, Wang_2022_CVPR} is a powerful tool for addressing this need by allowing the model parameters to change at the test time.
Nonetheless, much like the man who is no longer the same after stepping into the river, it is uncertain whether the adapted model will evolve in a better or worse direction.
For instance, under extended time horizons~\cite{hoang2024petta, press2023rdumb} or challenging testing streams~\cite{yuan2023robust, su2023realworld, gong2022note}, recent studies raise a critical concern that TTA models \textit{can be collapsed}, resulting in predictions that are confined to a single set of categories regardless of the input after several iterations.
As such, unforeseen risks are inevitable in TTA.
Deepening the understanding of these risks, including how they occur and how to prevent them, is crucial for ensuring a reliable, and trustworthy real-world TTA deployment.

While some risks may naturally arise during a TTA process~\cite{yuan2023robust, hoang2024petta, press2023rdumb, su2023realworld}, we delve into a more perilous situation where such systems are particularly vulnerable to malicious samples, which attackers could exploit to degrade the performance of a system intentionally.
This threat is known as \textit{adversarial model attack}~\cite{szegedy2014, goodfellow2015, yulong2023}.
To our best knowledge, only a limited number of prior studies investigated this threat on continual TTA models~\cite{wu2023uncovering, cong2024poisoning, park2024medbn}. 
Unfortunately, all resort to the white-box setting (assuming access to parameters of a victim model), which is impractical to implement.  
The gap here urges us to extend the concept of \textit{black-box}~\cite{nicolas2017_practical, guo19a_simple_black-box} attack to continual TTA that is not only more realistic but also a pioneering work in this area.

TTA updates a ML model using testing samples available at test time.
Thus, an attack algorithm functions by modifying that batch of testing samples~\cite{wu2023uncovering, cong2024poisoning, park2024medbn}. 
Theoretically speaking, the sampling process is manipulated. 
We aim to find a simple yet dangerous sampling operator, opening the ability to attack a TTA method effortlessly.
By extending Gaussian Mixture Model Classifier (GMMC)~\cite{hoang2024petta}, a handy theoretical model for understanding TTA, we first discover one such sampling strategy that can easily fool TTA to converge undesirably, under a condition. 
This condition is not hard to meet in almost every modern continual TTA methods~\cite{Wang_2022_CVPR, yuan2023robust, hoang2024petta, döbler2023robust} that requires a random image transformation~\cite{shorten2019survey} to be applied during training at test time.
The idea behind augmentation is straightforward, based on the fact that the semantics of an image are unchanged under mild random image transformations. 
These augmentation strategies generate more samples that favor the adaptation efficacy~\cite{hendrycks2021, pmlr-v124-lyzhov20a, marvin2022}.
Yet, this good practice becomes problematic when incorrectly predicted samples are augmented and used for adaptation. 
Ultimately, it turns into the \textit{``Achilles' heel"}\footnote{An idiom from \textit{Greek mythology} that refers to a weakness of a system or person that can lead to failure, despite the supreme overall strength.} of TTA that one can exploit to attack, even within the \textit{black-box} constraints.
Aside from the attack scenario, the discussion here is still relevant if a general user unintentionally queries samples that fall into this \textit{corner test case}.

This leads us to develop \textbf{\underline{R}eusing of \underline{I}ncorrect \underline{P}redictions (RIP)}, the first \textbf{black-box} attack algorithm to make a TTA model prone to collapse. RIP is illustrated by a simple binary classification task in \Fig{\ref{fig:overview}}. Here, the attacker intentionally picks incorrect predictions in previous TTA steps (highlighted in red) and reuses them in the subsequent testing batches. 
With random augmentation applied, incorrectly predicted samples and their augmented variants are used for TTA. 
We discover that under RIP, the decision boundary of a victim class is erroneously shifted, penetrated, and dominated by nearby classes.
Over time, a TTA model can be collapsed in this way.   
Undoubtedly, RIP is a straightforward attack algorithm that does not require any specialized expertise. The contributions of this work are:
\begin{itemize}
    \item Through a theoretical model, we \textit{discovered a threat} in which data augmentation in TTA and \textit{i.i.d.} sampling assumption violation can make a model collapse (\Sec{\ref{sec:theoretical}}).
    \item Inspired by this observation, \textit{Reusing of Incorrect Predictions (RIP)} - the first black-box attack algorithm targeting continual TTA methods is proposed (\Sec{\ref{sec:rip_attack}}).
    \item Through extensive experiments, we \textit{confirm the vulnerability} of many recent continual TTA methods (\Sec{\ref{sec:results}}). 
    \item A series of ablation studies\textit{ verifies the root causes} of vulnerability that can help to mitigate RIP attack (\Sec{\ref{sec:causes_of_risks}}).
\end{itemize}
Visit the Appendix for a summary of the related work. 

\section{The Continual TTA Procedure}
\label{sec:background}
This section presents the major notations and describes key components living in a continual TTA method.

\subsection{Continual Test-time Adaptation}
\noindent \textbf{Notations.} We focus on a ML classifier $f_t: \mathcal{X} \rightarrow \mathcal{Y}$, parameterized by $\theta_t \in \Theta$ (parameter space) that maps an input image $\boldsymbol{x} \in \mathcal{X}$ to a class-label $y \in \mathcal{Y}$. A TTA method continuously modifies $f_t$ at each time step $t \in \mathcal{T}$. 
Let the capital letters $(X_t, Y_t) \in \mathcal{X} \times \mathcal{Y}$ denote a pair of \textit{random variables} with the joint distribution $P_t(\boldsymbol{x},y) \in \mathcal{P}_d, t \in \mathcal{T}$. In practice, $X_t$ is in the form of a batch of $B$ testing samples. The superscript such as $X_t^{(i)}$ is used to denote the $i$-th realization of a random variable, when necessary.
The covariate shift~\cite{10.5555/1462129} is assumed: $P_t(\boldsymbol{x})$ and $P_{t'}(\boldsymbol{x})$ could be different but $P_t(y|\boldsymbol{x}) = P_{t'}(y|\boldsymbol{x})$ holds $\forall t \neq t'$.
At $t=0$, $\theta_0$ is a source model trained on labeled data from distribution $P_0$. 

\begin{figure}[t!]
    \centering
    \includegraphics[width=.95\linewidth]{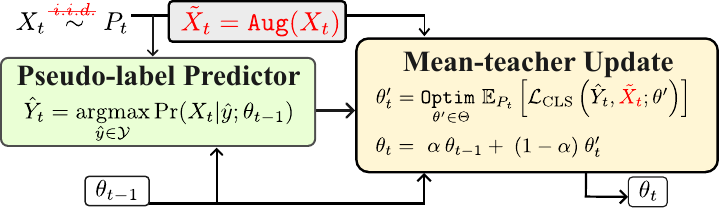}
    \vspace*{-0.5\baselineskip}
    \caption{The key operational steps of a continual TTA method. We extend the model in~\cite{hoang2024petta} (\textcolor{red}{red} highlighted) with the augmentation operator \texttt{Aug}($\cdot$) and investigate its effect in a vulnerable scenario where testing samples are \textit{not i.i.d.} sampled from a distribution $P_t$, but instead selectively sampled to collapse a model.}
    \label{fig:non-iid-model}
    \vspace*{-\baselineskip}
\end{figure}

\noindent \textbf{The Continual TTA Procedure. }
At $t>0$, the continual TTA process follows the diagram provided in \Fig{\ref{fig:non-iid-model}}:

\begin{enumerate}[label=(\roman*)]
    \item $X_t$ is sampled from distribution $P_t$. While \textit{i.i.d.} sampling is typically assumed, this work discusses a critical point: \textit{``What if this assumption fails to hold?"}

    \item The pseudo-label predictor (\Sec{\ref{sec:pseudo-label-predictor}}) guesses the label for $X_t$ based on $\theta_{t-1}$ (from the previous step):
    \begin{align}
    \label{eq:pseudo_label}
    \hat Y_t = f_{t-1}(X_t). %
    \end{align}
    \item Besides returning $\hat{Y}_t$ (\Eq{\ref{eq:pseudo_label}}) as a prediction, pseudo-labels are used for minimizing an objective function (\Sec{\ref{sec:tta_loss_functions}}), adapting itself $f_{t-1} \rightarrow f_t$ (\Sec{\ref{sec:tta_model_update}}). 
\end{enumerate}

\subsection{Loss Functions} 
\label{sec:tta_loss_functions}

 With only unlabeled data ($X_t$) available at test time, a \textit{pseudo label}~\cite{hyun2013_pseudo} ($\hat Y_t$ as in \Eq{\ref{eq:pseudo_label}}) is introduced for each $X_t$.
As a shorthand notation, we omit $t$  and denote the following \textit{probability vectors} $\boldsymbol{p}$ \textit{and} $\boldsymbol{q}$. Here, $p_{\hat{y}} = \Pr\{\hat{Y}_t = \hat{y} \}$ and $q_{\hat{y}} = \Pr\{f_t(X_t) = \hat{y}\}$, for $\hat{y} \in \mathcal{Y}$ represent the conditional probability of the pseudo-label predictor and the model assign label $\hat{y}$ for a given input sample
(i.e., the intermediate model output after softmax and before argmax). %
Existing loss functions in the field can be classified into two groups:

\noindent \textbf{Augmenting-free Loss Functions:} In the most basic form, pioneering studies~\cite{wang2021tent, niu2022efficient} adopt the \textit{Entropy (Ent)} loss:

\begin{align}
    \mathcal{L}_{\mathrm{Ent}}(\boldsymbol{q}) = - \sum_{\hat{y} \in \mathcal{Y}} q_{\hat{y}} \log\left( q_{\hat{y}} \right).
    \label{eq:ent_loss}
\end{align}

Utilizing the pseudo-label, the \textit{Cross Entropy (CE)}~\cite{NIPS2004_96f2b50b} loss replaces $\boldsymbol{p}$ in the position of one $\boldsymbol{q}$ in \Eq{\ref{eq:ent_loss}}: 
\begin{align}
    \mathcal{L}_{\mathrm{CE}}(\boldsymbol{p}, \boldsymbol{q}) = - \sum_{\hat y \in \mathcal{Y}} p_{\hat{y}} \log \left(q_{\hat{y}}\right),
    \label{eq:ce_loss}
\end{align}

While there are two replacement choices, a symmetric version - \textit{Symmetry Cross Entropy (SCE)} loss~\cite{marsden2024universal}:
\begin{align*}
    \mathcal{L}_{\mathrm{SCE}}(\boldsymbol{p}, \boldsymbol{q}) = -\frac{1}{2} \left(\sum_{\hat{y} \in \mathcal{Y}} p_{\hat{y}} \log q_{\hat{y}} + q_{\hat{y}} \log p_{\hat{y}} \right)
\end{align*}
is also commonly used.
In RMT~\cite{döbler2023robust}, an elaborated version that applies $\mathcal{L}_{\mathrm{SCE}}$ twice is used as their self-training loss ($\mathcal{L}_{\mathrm{RMT}}$).
To improve adaptation stability~\cite{marsden2024universal}, the \textit{Soft Likelihood Ratio (SLR)} loss~\cite{mummadi2021_slrloss} modifies CE as follows:
\begin{align}
    \mathcal{L}_{\mathrm{SLR}} (\boldsymbol{p}, \boldsymbol{q}) = - w \sum_{\hat{y} \in \mathcal{Y}} p_{\hat{y}} \log \left(\frac{q_{\hat{y}}}{\sum_{\hat{y}' \not = \hat{y}} q_{\hat{y}'}}\right),
    \label{eq:slr_loss}
\end{align}
with $w$ is the corresponding weight for sample $\hat y$.

\noindent \textbf{Augmenting Loss Functions: } Later studies~\cite{Wang_2022_CVPR, döbler2023robust, yuan2023robust,su2023realworld, marsden2024universal,nguyen2023tipi} further advance TTA with the use of augmented samples. 
With \texttt{Aug} is a random data augmentation operator, instead of using  $\boldsymbol{q}$ (e.g., in \Lss{CE} - \Eq{\ref{eq:ce_loss}}), $\boldsymbol{\Tilde{q}}$ is used with:
\begin{align}
    \Tilde{q}_{\hat{y}} = \Pr\{f_t\left(\Tilde{X}_t\right) = \hat{y}\}; \quad 
    \Tilde{X}_t = \text{\texttt{Aug}}\left(X_t\right).
    \label{eq:aug_pseudo_lbl}
\end{align}
Here, \texttt{Aug}$(X_t)$ replaces $X_t$. The consistency of the model output given $X_t$ and its diverse views $\Tilde{X}_t$, via random augmentation, is encouraged to increase TTA update efficacy.

\subsection{Model Update}
\label{sec:tta_model_update}
\noindent \textbf{TTA with Mean Teacher Update. } To achieve a stable optimization process, the main (\textit{teacher}) model $f_t$ are updated indirectly through a \textit{student} model $f_t'$ with parameter $\theta_t'$~\cite{wang2021tent, yuan2023robust, döbler2023robust,gong2022note, antti2017_mean_teachers}.
With $\mathcal{L}_{\mathrm{CLS}}$ as a placeholder for the loss function (\Sec{\ref{sec:tta_loss_functions}}), and a regularizer $\mathcal{R}$, the student model ($f_t'$) is first updated with a generic optimization operator \texttt{Optim}, followed by an \textit{Exponential Moving Average (EMA)} update of the teacher model parameter $\theta_{t-1}$:
\begin{align}
    \theta_t'&= \underset{\theta' \in \Theta}{\text{\texttt{Optim} }}  \mathbb{E}_{P_t}\left[\mathcal{L}_{\mathrm{CLS}} \left(\hat Y_t, X_t; \theta'\right) \right] + \lambda \mathcal{R}(\theta'), \label{eq:general_opti_step}\\ \ %
    \theta_t &= \alpha \theta_{t-1} + (1-\alpha)  \theta'_t ,\label{eq:general_teacher_update}
\end{align}
with $\alpha \in (0,1)$ - the EMA update rate, and $\lambda \in \mathbb{R}^+$ - the coefficient of the regularization term are hyper-parameters. 

\noindent \textbf{Source Model Weights Ensemble Update: } Suggested in ROID~\cite{marsden2024universal} to avoid self-training and mean teacher update~\cite{antti2017_mean_teachers}. The accumulated model is updated via: 
    \begin{align}
        \theta_t &=\alpha \theta_{0} + (1-\alpha) \theta'_t,
        \label{eq:source_weights_ensemble_update}
    \end{align}
with $\theta_0$ is the parameter of the source model.

\subsection{Pseudo-label Predictor}
\label{sec:pseudo-label-predictor}
The pseudo-label of augmented samples ($\hat{y}_t$) in \Eq{\ref{eq:aug_pseudo_lbl}} can be predicted in two ways. Let denote $\bar{f}_t$ be the model updated with EMA in \Eq{\ref{eq:general_teacher_update}} (teacher model). We have:

\begin{itemize}
    \item \textbf{Pseudo-labels from Teacher Model:} The earliest work - CoTTA~\cite{Wang_2022_CVPR} proposes the following strategy: 
    \begin{align}
        \hat{y}_t = \bar{f}_{t-1}(\Tilde{x}_t).
        \label{eq:pseudo_from_teacher}
    \end{align}
    \item \textbf{Pseudo-labels from Student Model: } This strategy is used in almost every follow-up study after CoTTA, (e.g., RMT~\cite{döbler2023robust}, ROID~\cite{marsden2024universal}, RoTTA~\cite{yuan2023robust} or TRIBE~\cite{su2023realworld}):
\begin{align}
    \hat{y}_t = f_{t-1}(\Tilde{x}_t).
    \label{eq:pseudo_from_student}
\end{align}
\end{itemize}

\section{A Risk in TTA with Data Augmentation}
\label{sec:theoretical}

\begin{figure*}[ht!]
    \centering
    \includegraphics[width=.95\linewidth]{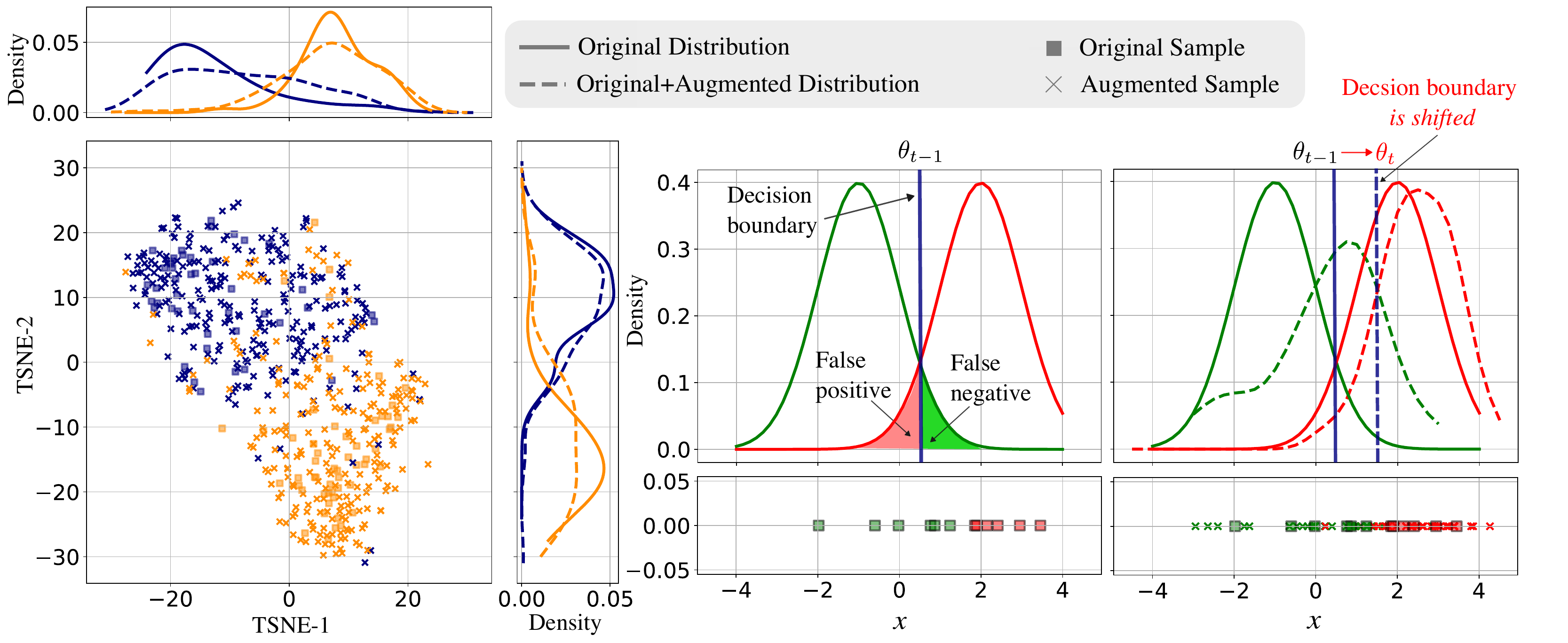}
    \vspace*{-0.5\baselineskip}
    \caption{
    The similarity in the effect of random data augmentation on images of  CIFAR-10-C~\cite{hendrycks2019robustness} and synthetic data used in Gaussian Mixture Model Classifier (GMMC).
    (left)
    2D \textit{t-SNE}~\cite{maaten2008_tsne}  projection of the deep feature vectors of real images.
    (middle) Samples drawn from two Gaussian distributions and the best theoretical separation boundary on GMMC (solid line). Regions with incorrect predictions are highlighted.   
    (right) An Additive White Gaussian Noise (AWGN) generates augmented samples for GMMC and the decision boundary separates original and augmented samples (dashed line). 
    The shifting of decision boundaries when training with augmented samples is similarly observed on both real images and GMMC simulated data, allowing our analysis to focus on GMMC+AWGN for their simplicity.
    }
    \label{fig:t-SNE}
    \vspace*{-0.5\baselineskip}
\end{figure*}

By inspecting a toy example of a simple theoretical model (\Sec{\ref{sec:gmmc_analysis}}),
this section delves into our first findings on a sampling strategy, namely \textit{Incorrectly Prediction Sampling} (IPS) (\Sec{\ref{sec:ips_gmmc}}) that negatively impacts a TTA method.

\subsection{Augmented Gaussian Mixture Model Classifier}
\label{sec:gmmc_analysis}

In~\cite{hoang2024petta}, Hoang \textit{et al.} introduces a simple yet representative Gaussian Mixture Model Classifier (GMMC) that replicates the behavior of a real-world continual TTA model for a theoretical analysis. 
However, \textit{GMMC fails to account for the role of the random image augmentation operator, }a common practice in modern continual TTA. 
This study extends GMMC along this line and introduces an inspection, shedding light on the introduction of our attack algorithm.

To replicate the \texttt{Aug} operator in \Eq{\ref{eq:aug_pseudo_lbl}} on GMMC, we employ \textit{Additive White Gaussian Noise (AWGN)} that can be directly applied to the data used for optimizing GMMC: 
\begin{align}
    \Tilde{X}_t = 
    \mathrm{Aug}_\sigma(X_t) = X_t + \delta, 
    \qquad \delta \sim \mathcal{N}(0, \sigma)     
    \label{eq:gmmc_aug}.
\end{align}
The level of data augmentation can be controlled by $\sigma$. 
A larger value of $\sigma$ corresponds to a stronger augmentation scheme, and $\sigma = 0$ means no data augmentation is applied.

While AWGN is relatively simple, \Fig{\ref{fig:t-SNE}} empirically justifies its validity as a \textit{proxy for image-based operators} \texttt{Aug}. 
In \Fig{\ref{fig:t-SNE}}-(left), the 2D \textit{t-SNE}~\cite{maaten2008_tsne} projects deep-feature embedding (source model) of 100 
random CIFAR-10-C~\cite{hendrycks2019robustness} images from two classes. Five augmented variations (using common augmentation operators - see the Appendix) are visualized together with the original images. 
Data augmentation expands the coverage of a sample into multiple directions around their initial data points (similar to findings in~\cite{madry2018towards, zhang2018mixup}).  
This effect is highly analogous to the one introduced by AWGN on 1-dimensional data used in GMMC. 
In sum, GMMC with AWGN in \Eq{\ref{eq:gmmc_aug}} \textit{can potentially serve as a surrogate model for analyzing a TTA model with image-augmentation operator} introduced in~\Eq{\ref{eq:aug_pseudo_lbl}}.

\subsection{Incorrect Prediction Sampling on GMMC}
\label{sec:ips_gmmc}

\begin{figure}[ht!]
    \centering
    \includegraphics[width=.85\linewidth]{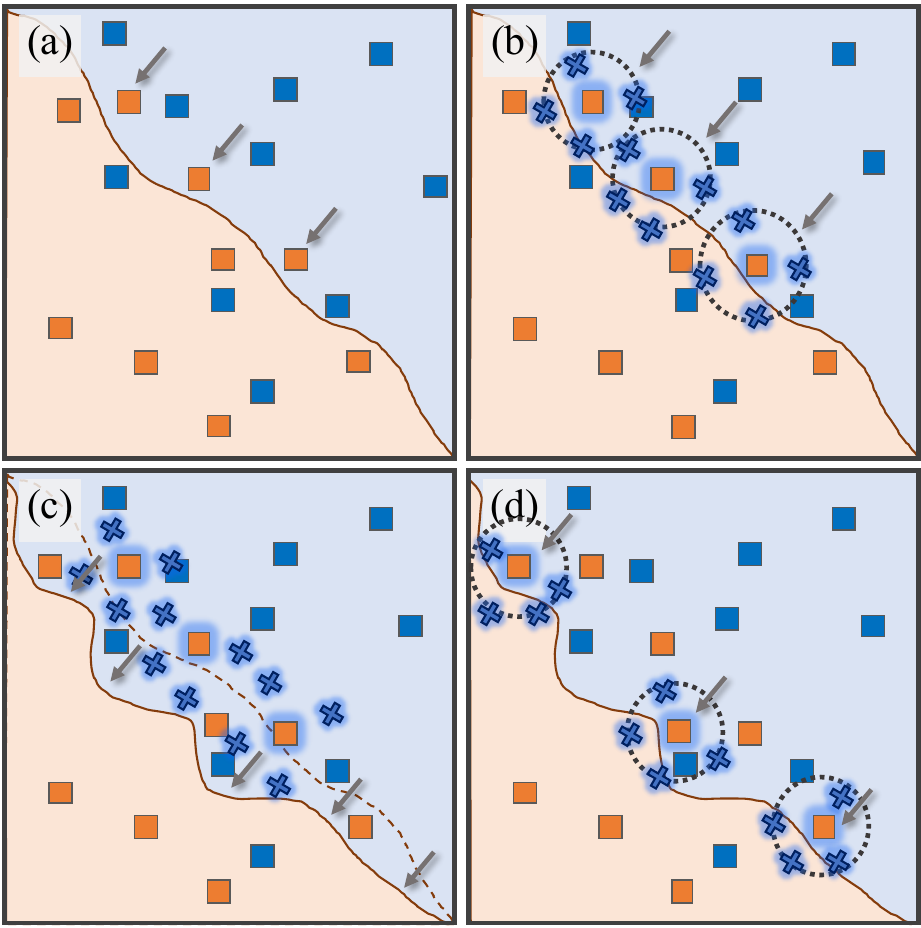}
    \vspace*{-0.5\baselineskip}
    \caption{A step-by-step illustration of the shifting-boundary effect caused by incorrect prediction sampling (IPS), elaborating on \Fig{\ref{fig:t-SNE}}-left. Arrows serve as pointers. 
    (a) Mispredicted samples from a victim class (orange)  are sampled for TTA. 
    (b) Randomly augmented variations (denoted by $\times$) are generated, expanding the area (dashed circles) around the original samples. 
    (c) The updated decision boundary expands to cover these samples - highlighted with the blue halo effect, penetrating the victim class. 
    (d) The process repeats, reducing the chance of predicting the victim class. 
    \vspace*{-1.5\baselineskip}
    }
    \label{fig:rip-cartoon}
\end{figure}

\begin{figure*}[t!]
    \begin{subfigure}[b]{0.24\textwidth}
    \centering
    \includegraphics[width=1\linewidth]{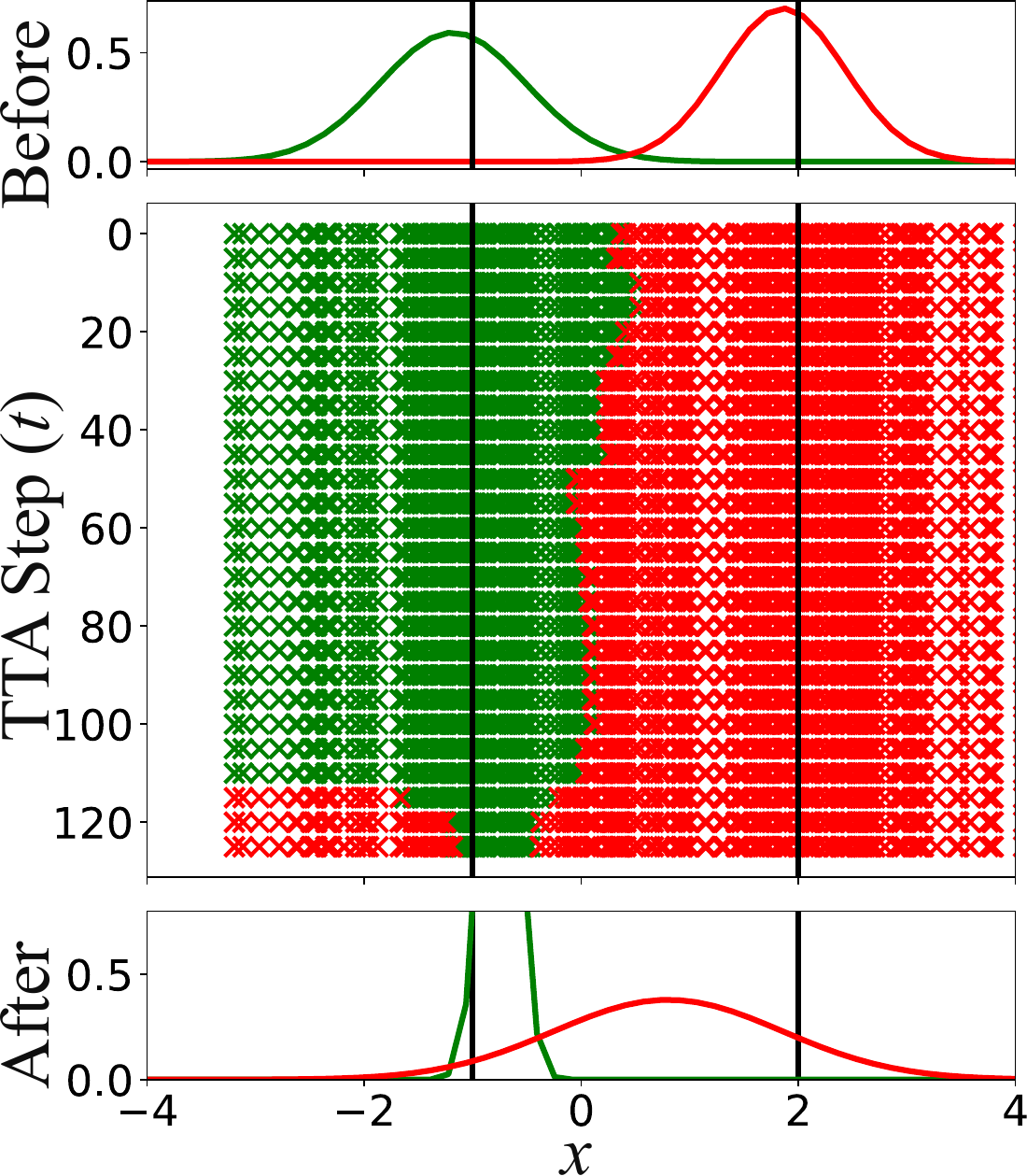}
    \caption{IPS: \cmark, \texttt{Aug}: \cmark, $\alpha=0.9$.}
    \label{fig:simulation_collapsed}
    \end{subfigure}%
    \begin{subfigure}[b]{0.24\textwidth}
    \includegraphics[width=1\linewidth]{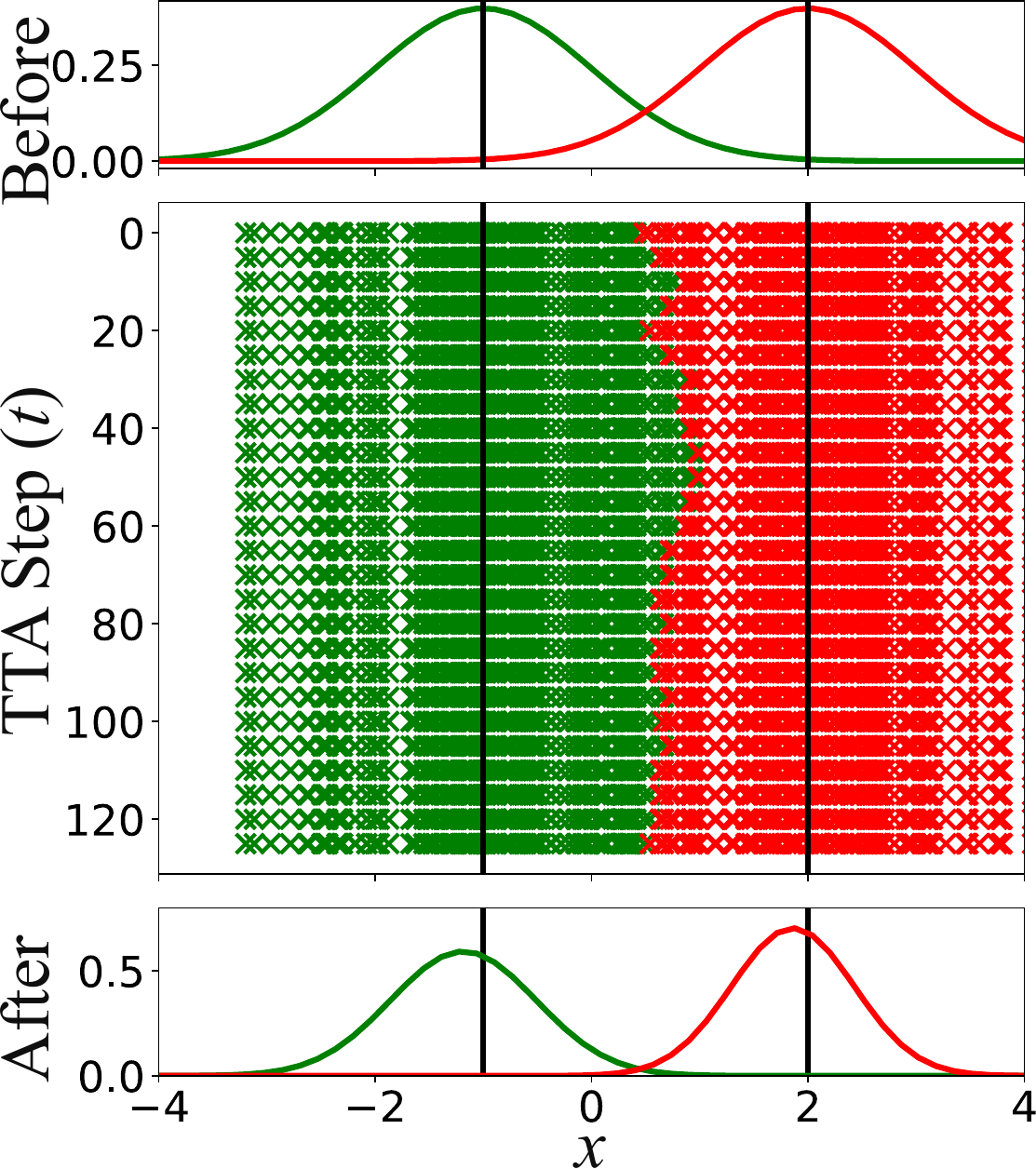}
    \caption{IPS: \xmark, \texttt{Aug}: \cmark, $\alpha=0.9$.}
    \label{fig:simulation_none_collapsed}
    \end{subfigure}%
    \begin{subfigure}[b]{0.24\textwidth}
    \includegraphics[width=1\linewidth]{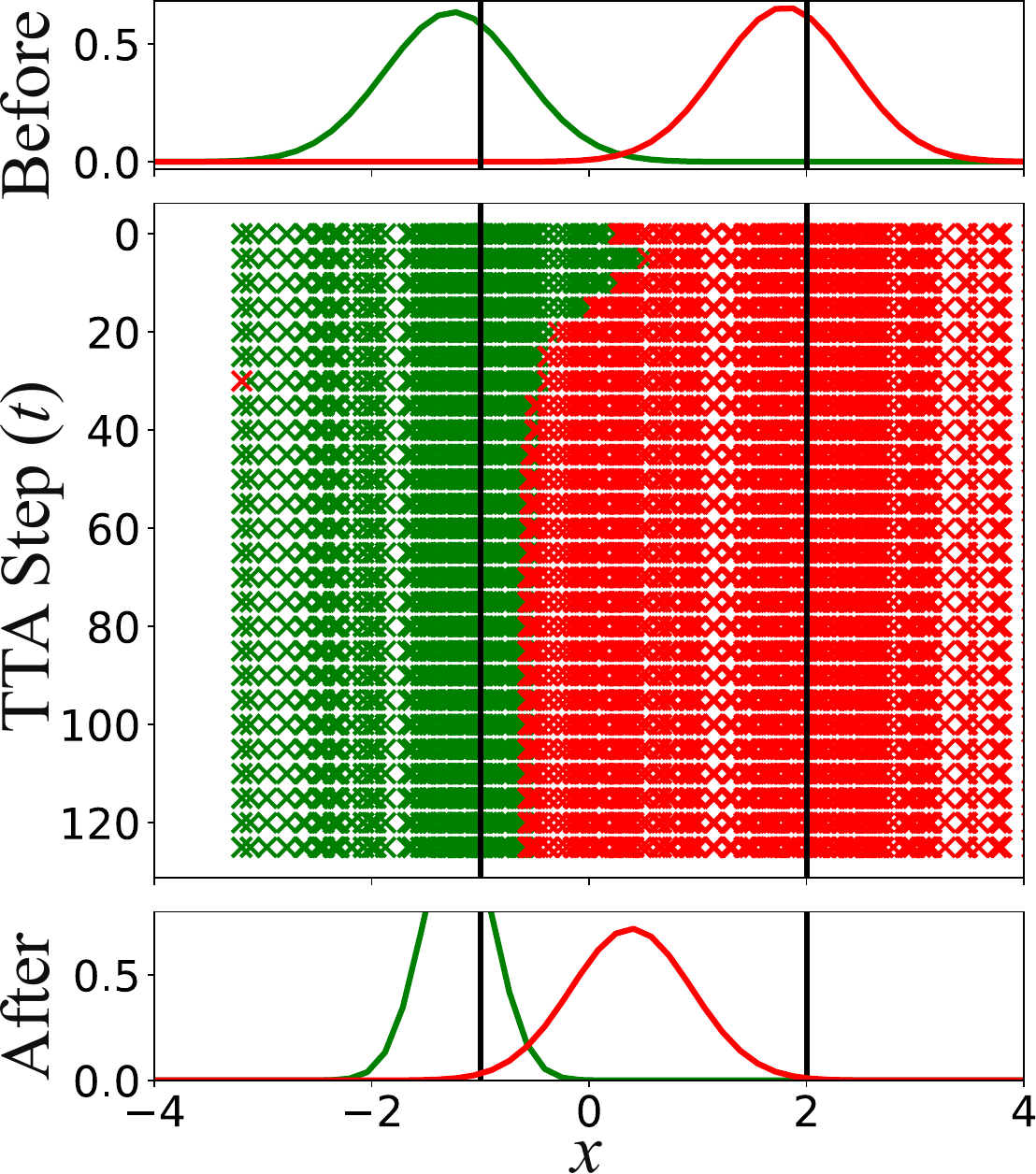}
    \caption{IPS: \cmark, \texttt{Aug}: \xmark, $\alpha=0.9$.}
    \label{fig:simulation_no_aug}
    \end{subfigure}%
    \begin{subfigure}[b]{0.24\textwidth}
    \includegraphics[width=1\linewidth]{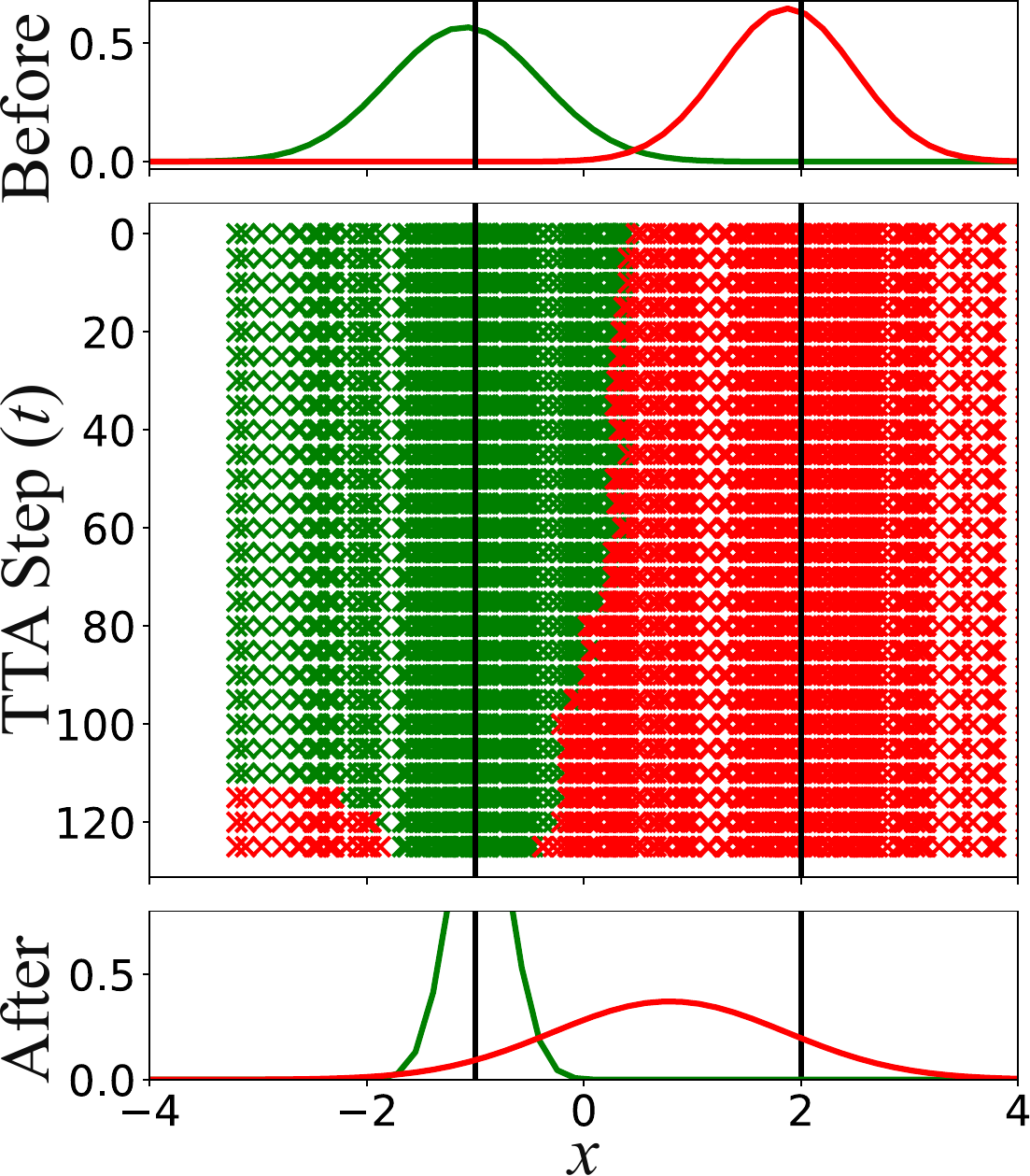}
    \caption{IPS: \cmark, \texttt{Aug}: \cmark, $\alpha=0.95$.}
    \label{fig:gmmc_alpha95-1}
    \end{subfigure}
    \vspace*{-0.5\baselineskip}
    \caption{Simulation results on the Gaussian Mixture Model Classifier (GMMC) representing the effect of Incorrect Prediction Sampling (IPS), augmentation operator (\texttt{Aug}), and update rate ($\alpha$).
    (\cmark) denotes if the operator is enabled, and (\xmark) vice versa.
    The distribution before (top) and after adaptation (bottom) is visualized. The (middle) plot shows the shifting in model prediction (on the same set of samples) after every 20 steps.
    (a)-(c) GMMC is collapsed if IPS and \texttt{Aug} are simultaneously enabled. 
    (d) Increasing $\alpha$ partially mitigates the collapse.  
    }
    \label{fig:simulation_gmmc}
    
\end{figure*}

\noindent \textbf{The Shifting-Boundary Effect. }
Since random augmentation can generate variations around an original sample, the optimal decision boundary optimized on a combination of augmented and original samples \textit{can be shifted} as shown in \Fig{\ref{fig:t-SNE}}-right.
It is noted that the distribution of the original did not change, and augmented samples near the decision boundary make this drift. 
Finding samples in this area is straightforward as they are likely to be incorrectly predicted, either false positives or negatives (\Fig{\ref{fig:t-SNE}}-middle).

\noindent \textbf{Incorrect Prediction Sampling (IPS)}. The shifting boundary effect gives an idea of \textit{a sampling operator} that $X_t$ is only selected if it comes from a victim class ($y_a$) and its pseudo-label is incorrectly predicted (i.e., $\hat{Y}_t \neq Y_t \land Y_t = y_a$).
We name this strategy as \textit{IPS}. 
When $X_t$'s are not \textit{i.i.d.} sampled from $P_t$, but IPS operator is applied instead, an attacker can easily modify the decision boundary of a TTA model. 
\Fig{\ref{fig:rip-cartoon}} illustrates the shifting boundary effect caused by IPS. 
Due to the expansion of incorrect predictions of the victim class via augmentation and TTA model update, its decision boundary is penetrated by the nearby classes.

\noindent \textbf{Numerical Simulation. }
To empirically confirm the case of IPS and \texttt{Aug} on GMMC, we carry out a numerical simulation in \Fig{\ref{fig:simulation_gmmc}}, with $T=120$ adaptation steps (See the Appendix for setup details).
The collapse is observed when most predictions converge to a single label, no matter what the input data is (see Def.~\ref{def:model_collapse}).
In \Fig{\ref{fig:simulation_collapsed}}, the model is collapsed when IPS is applied while this is not the case in \Fig{\ref{fig:simulation_none_collapsed}}. 
When removing \texttt{Aug} in \Fig{\ref{fig:simulation_no_aug}}, the boundary can be shifted, but a total collapse is not observed.
In sum, the collapse only happens when the \textit{two conditions are met}: IPS is performed and \texttt{Aug} operator is in place. 
Slowing down the update $\alpha$ (\Eq{\ref{eq:general_teacher_update}}) helps mitigate the effect of RIP (\Fig{\ref{fig:gmmc_alpha95-1}}). We will return back to this discussion in \Sec{\ref{sec:study_update_schemes}}.

\section{Reusing Incorrect Predictions (RIP) Attack}
\label{sec:rip_attack}
Inspired by IPS, this section establishes the threat model and our RIP attack, an algorithm introducing this threat.
\begin{table*}[t!]
    \centering
    \resizebox{.95\linewidth}{!}{ 
    \begin{tabular}{m{1.5cm}|m{2.0cm}|m{6.5cm}|m{1.6cm}|m{1.6cm}|m{1.5cm}|m{1.0cm}}
         \toprule
         \textbf{Attack Type} & \multicolumn{2}{l|}{\textbf{Attack Objective}} & \textbf{Attack \newline Algorithm} &
         \textbf{Model \newline Parameters} &
         \textbf{Benign Sample} & 
         \textbf{Source Dataset}
         \\
         \midrule
         
         \multirow{3}{*}{White-box} &
         \cellcolor{gray!40}Targeted & 
         \cellcolor{gray!40} Flipping the prediction of a victim category (class) to a target category&
         
         \multirow{1}{1.5cm}{TePA~\cite{cong2024poisoning}}&
         \multirow{1}{*}{\cmark} & 
         \cmark & 
         \cmark
         \\
         & 
         Stealthy \newline Targeted & 
         Achieving the above objective while maintaining the performance in other categories&
         \multirow{2}{1.5cm}{DIA~\cite{wu2023uncovering}}&
         \multirow{2}{*}{\cmark} &
         \multirow{2}{*}{\cmark} & 
         \multirow{2}{*}{\xmark}
         \\
         
         & 
         \cellcolor{gray!40} Indiscriminate & 
         \cellcolor{gray!40} Degrading the performance of all categories &
         &
         &
         \\
         \midrule
         Black-box \textit{(ours)}& 
         Collapsing \textit{(ours)} & 
         Degrading the performance of at least one (or more) category&
         RIP \textit{(ours)} &
         \xmark &
         \xmark & 
         \xmark
         \\
        \bottomrule
    \end{tabular}
    }
    \vspace*{-0.3\baselineskip}
    \caption{
    A comparison of existing white-box TTA attack attempts in~\cite{wu2023uncovering, cong2024poisoning} and \textbf{RIP}-our proposed \textbf{black-box} attack algorithm that makes a TTA model \textbf{collapse}. 
    A summary of the main attack objectives and the representative attack algorithms are provided.  
    For each algorithm, we highlight whether an assumption is necessary (\cmark) or not (\xmark) for a success attack. 
    The assumptions are: 
    (Model Parameters) - accessing model parameters at any time, either before (source model) or after an adaptation step; 
    (Benign Sample) - intercepting and pixel-level modifying benign samples from other users to generate malicious samples;
    (Source Dataset) - accessing or sampling from the distribution that creates the source dataset.
    Although the attack goal is weaker than the ones in the white-box attack, our \textit{collapsing attack} is still considered dangerous in practice.
    This goal is realized by RIP - an attack algorithm that lifts all assumptions made by previous attempts. 
    }
    \label{tab:attack_goals}
    \vspace*{-\baselineskip}
\end{table*}

\subsection{Threat Model}
\label{sec:collapsing_attack}
The threat model in this study focuses on two main aspects: first, making a TTA \textit{model collapsed} as the main objective and second, \textit{black-box} as the constraint for the attack. 

\noindent \textbf{Collapsing Attack Objective.}
\textit{Collapsing attack on continual TTA is first introduced in this study}.
Its primary objective is to make a continual TTA model that tends to \textit{ignore} some categories in $\Tilde{\mathcal{Y}} \subset \mathcal{Y}$. This is called model collapse, and Definition~\ref{def:model_collapse} restates its mathematical definition in~\cite{hoang2024petta}: 
\begin{definition}[\textbf{Model Collapse}]
\label{def:model_collapse}
A model is said to be \textit{collapsed} from step $\tau \in \mathcal{T}, \tau < \infty$ if there exists a non-empty subset of categories $\Tilde{\mathcal{Y}} \subset \mathcal{Y}$ such that $\Pr\{Y_t \in \Tilde{\mathcal{Y}}\} > 0$ but the marginal $\Pr\{\hat Y_t \in \mathcal{\Tilde{Y}}\}$ converges to zero in probability: 
\begin{align*}
    \underset{t \to \tau}{\lim} \Pr\{\hat Y_t \in \Tilde{\mathcal{Y}}\} = 0. %
\end{align*}%
\end{definition}

As it is irrecoverable once collapsed, the only remedy would be resetting all parameters back to $\theta_0$. \Tab{\ref{tab:attack_goals}} compares the existing continual TTA attack objectives~\cite{wu2023uncovering} and our TTA model collapsing TTA attack (columns 2-3).

\noindent \textbf{Metric for Collapsing Attack. } We simply compute the average class-wise testing error among all categories and report the increment versus the same model without attack.

\noindent \textbf{Black-box Continual TTA Model Attack. } In a TTA black-box attack, an attacker can only interact with queries and model responses~\cite{nicolas2017_practical}.
Before and during the adaptation, accessing the model architecture and parameters, the operating TTA algorithm, gradient information, optimizer state, etc., or queries from other users are strictly forbidden at any time. We define the terminology of \textit{black-box TTA attack} as any attack algorithm following these constraints. 

\subsection{Reusing Incorrect Prediction Attack Algorithm}
\label{sec:attack_description}
\begin{algorithm}[ht!] %
    \LinesNumbered
    \KwIn{Labeled dataset $\mathcal{D}_a$ for attack, victim attack class $y_a$, victim continual TTA model $f_t(x)$, testing batch size $B$, the number of attack rounds $T_a$.} 

     \tcp{Initialize $\mathcal{S}_0$ with samples in $\mathcal{D}_a$}
    $\mathcal{S}_0 \leftarrow \left\{ \left(X_0^{(i)}, Y_0^{(i)}\right) \sim \mathcal{D}_a  \right\}_{i=1}^{B}$
    
    \For{$t \in [1, \cdots, T_a]$}{ 
        
        \tcp{Predictions from TTA model}
        $\hat Y_{t-1}^{(i)} \leftarrow f_{t-1}\left(X_{t-1}^{(i)}\right), X_t^{(i)} \in \mathcal{S}_{t-1}$

        \tcp{Set of incorrect predictions}
        $\mathcal{I}_t \leftarrow \left\{ X_t^{(i)} | Y_t^{(i)} \not = \hat Y_t^{(i)} \land Y_t^{(i)} = y_a, i=1 \dots B \right\}$

        \tcp{Fulfilling with samples from $\mathcal{D}_a$}
        
        $\mathcal{S}_t \leftarrow \mathcal{I}_t \cup \left\{ \left(X_t^{(i)}, Y_t^{(i)}\right) \sim \mathcal{D}_a \right\}_{i=1}^{B-|\mathcal{I}_t|}$
    }
    \caption{\underline{R}eusing \underline{I}ncorrect \underline{P}redictions (RIP) Attack}
    \label{alg:rip_attack}
\end{algorithm} %
\vspace*{-0.5\baselineskip}

\noindent \textbf{Attack Description.}
The idea of \textit{\underline{R}eusing \underline{I}ncorrect \underline{P}rediction (RIP)} is as simple as IPS. 
RIP capitalizes on the vulnerability of TTA in \Sec{\ref{sec:theoretical}} by intentionally reusing incorrect predictions from one victim class $y_a$ in subsequent adaptation steps.
All mispredicted samples from $y_a$ in the previous steps are accumulated and reused.
A severe domain shift and the imperfection of $f_{t-1}$ under this new distribution make the pseudo-label in \Eq{\ref{eq:pseudo_label}} erroneous.
Hence, finding incorrect predictions is convenient. 
The only requirement here for RIP is that the attacker has access to a labeled dataset ($\mathcal{D}_a$) that is reasonably large to find at most $B$ incorrect predictions.  
\textit{RIP strictly follows the black-box setting} and does not add any specific or unusual setup that favors the attackers over the victim's continual TTA model. \Alg{\ref{alg:rip_attack}} gives the pseudo-code and \Fig{\ref{fig:overview}} provides a graphical illustration of our RIP attack algorithm in a simple case. 

\subsection{Comparison to Prior TTA Attack Studies}
\label{sec:comparision}

\noindent \textbf{Black-box versus White-Box Attack.} White-box continual TTA attacks assume the adversary has complete knowledge of the underlying TTA model.
Poisoning~\cite{cong2024poisoning} or distribution invading~\cite{wu2023uncovering, park2024medbn} attacks (DIA) fall into this category, where the source model (or the one before adaptation) is used as a surrogate model for generating adversarial attack samples.
Simply studying the developer manual of a popular ML API, for instance, ones from Open AI\footnote{\url{https://platform.openai.com/docs/api-reference}}, the white-box attack is obviously unrealistic to implement.
Most recent APIs \textit{do not reveal the source model or dataset it was trained on to general users.}
Hence, a black-box attack is the only possibility. 
\Tab{\ref{tab:attack_goals}} summarizes a comparison between white- and black-box algorithms (columns 4-7).

\noindent \textbf{RIP versus Other Attack Algorithms. } RIP is the first black-box attack attempt that stands out from existing ones. 
We note that previous algorithms generate adversarial samples by directly adding pixel-level perturbations to the original images. While these subtle changes may be \textit{imperceptible to humans, a trained adversarial detector can identify and reject malicious samples}~\cite{liang2021detecting, aldahdooh2022adversarial, tianyu2018, yulong2023} generated.

\section{Continual TTA Methods Under RIP Attack}
\label{sec:results}
\subsection{Experimental Setup}

\noindent \textbf{Continual TTA Task, Dataset and Methods.}
The effect of RIP attack is evaluated on the image classification task, with three benchmarks including CIFAR10 $\rightarrow$ CIFAR10-C, CIFAR100 $\rightarrow$ CIFAR100-C, and ImageNet $\rightarrow$ ImageNet-C~\cite{hendrycks2019robustness}.
The following continual TTA methods ($f_t$ in \Alg{\ref{alg:rip_attack}}) are studied: CoTTA~\cite{Wang_2022_CVPR}, EATA~\cite{niu2022efficient}, RMT~\cite{döbler2023robust}, RoTTA~\cite{yuan2023robust}, ROID~\cite{marsden2024universal}, TRIBE~\cite{su2023realworld}, and PeTTA~\cite{hoang2024petta}. 

\noindent \textbf{Attack Scenario.} We employ RIP attack follows \Alg{\ref{alg:rip_attack}}, with $T_a=500$ rounds, $B=64$ and $\mathcal{D}_a$ is a set of images corrupted by impulse noise from each dataset. 
The choice of corruption here is arbitrary and $B$ follows prior studies. Results in the Appendix show that other options also perform well.
In following experiments, we compute the error of each class independently, after every $25$ adaptation step, and report the average value among all classes.
For a robust estimation, we repeat the attack $10$ times (trials), each with a different victim attack label $y_a$, randomly selected and averaged across trials.
For all compared TTA methods, we use the default set of hyper-parameters from their authors.

\subsection{Vulnerability of Existing TTA Methods}
\label{sec:results_existingTTA}
\begin{table*}[ht!]
    \centering
    \resizebox{.95\textwidth}{!}{
    \begin{tabular}{l|c|cc|cc|cc}
\hline
\multirow{2}{*}{\textbf{Method}} & \multirow{2}{*}{\textbf{Venue}} & \multicolumn{2}{c|}{\textbf{CIFAR-10-C}}                       & \multicolumn{2}{c|}{\textbf{CIFAR-100-C}}                      & \multicolumn{2}{c}{\textbf{ImageNet-C}}                       \\ \cline{3-8} 
                                 &  & \textbf{No Attack}  & \textbf{RIP Attack} & \textbf{No Attack}  & \textbf{RIP Attack} & \textbf{No Attack}  & \textbf{RIP Attack} 
\\ \midrule
                          No TTA &          - &   0.7292 &  0.7292\inc{0} &    0.3937 &  0.3937\inc{0} & 0.8155 &  0.8155\inc{0} \\ \midrule
     PeTTA~\cite{hoang2024petta} & NeurIPS'24 &   0.3385 & 0.4704\inc{39} &    0.3634 & 0.4617\inc{27} & 0.7439 & 0.8174\inc{10} \\
     RoTTA~\cite{yuan2023robust} &    CVPR'23 &   0.4492 & 0.5872\inc{31} &    0.3751 & 0.5354\inc{43} & 0.8105 &  0.8237\inc{2} \\
ROID~\cite{marsden2024universal} &    WACV'24 &   0.2836 & 0.3719\inc{31} &    0.2885 & 0.4729\inc{64} & 0.7061 &  0.7536\inc{7} \\
    TRIBE~\cite{su2023realworld} &    AAAI'24 &   0.4537 & 0.5502\inc{21} &    0.3608 & 0.5691\inc{58} & 0.7737 &  0.7978\inc{3} \\
     CoTTA~\cite{Wang_2022_CVPR} &    CVPR'22 &   0.3105 & 0.3417\inc{10} &    0.4262 &  0.4383\inc{3} & 0.7524 &  0.7593\inc{1} \\
     RMT~\cite{döbler2023robust} &    CVPR'23 &   0.2871 & 0.3502\inc{22} &    0.3371 & 0.4155\inc{23} & 0.7182 &  0.7496\inc{4} \\
    EATA~\cite{niu2022efficient} &    ICML'22 &   0.2952 & 0.3488\inc{18} &    0.2926 & 0.4251\inc{45} & 0.7568 &  0.7571\inc{0} \\
    \bottomrule
    \end{tabular}}
    \caption{Average of the class-wise testing error (lower is better) across $500$ adaptation steps and $10$ RIP attack trials of the studied TTA methods~\cite{hoang2024petta, yuan2023robust, marsden2024universal, su2023realworld, Wang_2022_CVPR, döbler2023robust, niu2022efficient}. The performance of the source model (No TTA) and the TTA method without RIP attack (No Attack) are included for comparison.  
    While the damage caused by RIP may vary across algorithms, most state-of-the-art continual TTA approaches are severely affected. 
    The number in brackets shows the percentage increase in testing error under the RIP attack compared to no attack.}
    \label{tab:rip_attack_compared_methods}
    \vspace*{-.5\baselineskip}
\end{table*}

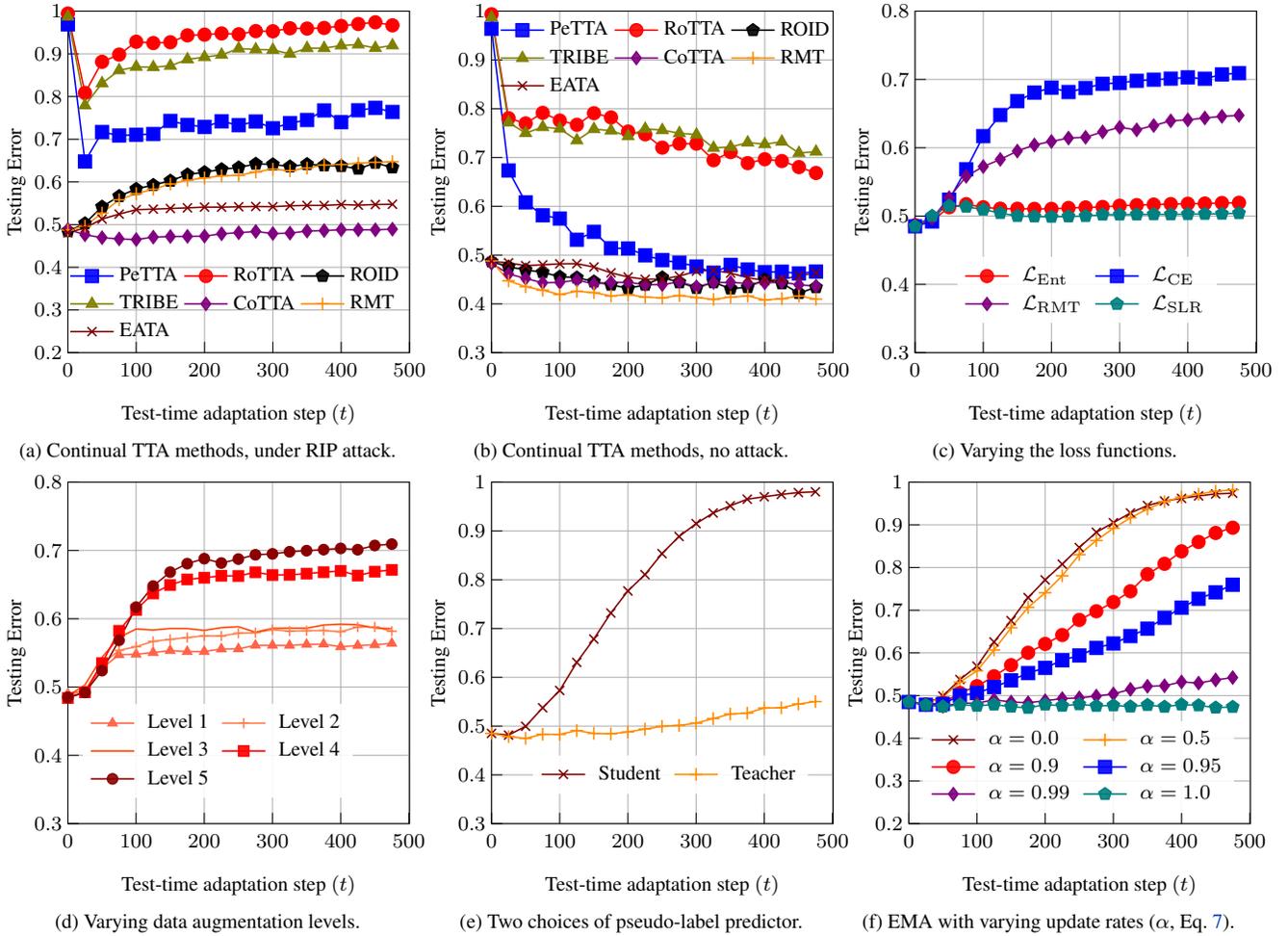
\begin{figure*}[ht!]
    \begin{subfigure}[b]{0.33\textwidth} %
        \centering
        \begin{tikzpicture}
\definecolor{darkgray176}{RGB}{176,176,176}
\definecolor{darkorange}{RGB}{255,140,0}
\definecolor{green01270}{RGB}{0,127,0}
\definecolor{lightgray204}{RGB}{204,204,204}
\definecolor{maroon}{RGB}{128,0,0}
\definecolor{olive}{RGB}{128,128,0}
\definecolor{purple}{RGB}{128,0,128}
\tikzstyle{every node}=[font=\footnotesize]
\begin{axis}[
legend cell align={left},
legend style={
  fill=none,
  fill opacity=0.9,
  draw opacity=1,
  text opacity=1,
  at={(-0.04,0.28)},
  anchor=north west,
  draw=none,
  column sep = 0,
  legend columns=3,
  font=\footnotesize
},
xtick distance=100,
ytick distance=0.1,
xmajorgrids,
ymajorgrids,
tick align=outside,
tick pos=left,
x grid style={darkgray176},
xlabel={Test-time adaptation step $(t)$},
xmin=0, xmax=500,
xtick style={color=black},
y grid style={darkgray176},
ylabel={Testing Error},
xtick align=inside,
ytick align=inside,
xticklabel style = {yshift=-1.0pt, font=\footnotesize},
yticklabel style = {xshift=4pt, , font=\footnotesize},
height=6.3 cm, width = 6.3 cm,
ymin=0.2, ymax=1.0,
ytick style={color=black},
outer sep=0pt,
outer xsep=3pt,
every axis x label/.style={
    at={(axis description cs:0.5,-0.18)},
    align = center},
every axis y label/.style={
    rotate=90,
    at={(axis description cs:-0.15,0.5)},
    align = center},
]
\addplot [semithick, blue, mark=square*, mark size=2.5, mark options={solid}]
table {%
0 0.969
25 0.6478
50 0.7168
75 0.7088
100 0.7103
125 0.7121
150 0.7429
175 0.7334
200 0.7289
225 0.7418
250 0.7328
275 0.7414
300 0.7256
325 0.7376
350 0.7451
375 0.7675
400 0.7401
425 0.7678
450 0.7735
475 0.7638
};
\addlegendentry{PeTTA}
\addplot [semithick, red, mark=*, mark size=2.5, mark options={solid}]
table {%
0 0.9944
25 0.8085
50 0.8814
75 0.8983
100 0.9284
125 0.9253
150 0.9272
175 0.9433
200 0.945
225 0.9478
250 0.9459
275 0.9533
300 0.9529
325 0.9599
350 0.9592
375 0.9607
400 0.9653
425 0.9699
450 0.9739
475 0.9668
};
\addlegendentry{RoTTA}
\addplot [semithick, black, mark=pentagon*, mark size=2.5, mark options={solid}]
table {%
0 0.4847
25 0.5028
50 0.5418
75 0.5666
100 0.5834
125 0.5926
150 0.6026
175 0.6172
200 0.6226
225 0.6294
250 0.6336
275 0.6422
300 0.6409
325 0.6359
350 0.6415
375 0.6385
400 0.637
425 0.6308
450 0.6444
475 0.6336
};
\addlegendentry{ROID}
\addplot [semithick, olive, mark=triangle*, mark size=2.5, mark options={solid}]
table {%
0 0.9872
25 0.7793
50 0.831
75 0.8615
100 0.8696
125 0.8687
150 0.872
175 0.8871
200 0.8924
225 0.8978
250 0.9124
275 0.9104
300 0.9093
325 0.9
350 0.9133
375 0.9134
400 0.9196
425 0.9218
450 0.9136
475 0.9198
};
\addlegendentry{TRIBE}
\addplot [semithick, purple, mark=diamond*, mark size=2.5, mark options={solid}]
table {%
0 0.489
25 0.4762
50 0.4694
75 0.4665
100 0.465
125 0.4706
150 0.4721
175 0.4724
200 0.473
225 0.4785
250 0.481
275 0.4839
300 0.479
325 0.48
350 0.4849
375 0.486
400 0.488
425 0.4879
450 0.4878
475 0.4893
};
\addlegendentry{CoTTA}
\addplot [semithick, darkorange, mark=+, mark size=2.5, mark options={solid}]
table {%
0 0.4874
25 0.4964
50 0.5269
75 0.5581
100 0.5724
125 0.5829
150 0.5951
175 0.6041
200 0.6092
225 0.6142
250 0.6153
275 0.6232
300 0.6299
325 0.6262
350 0.6325
375 0.6394
400 0.6408
425 0.6436
450 0.6459
475 0.6474
};
\addlegendentry{RMT}
\addplot [semithick, maroon, mark=x, mark size=2.5, mark options={solid}]
table {%
0 0.481
25 0.4922
50 0.513
75 0.5246
100 0.535
125 0.5362
150 0.5377
175 0.5392
200 0.541
225 0.5408
250 0.5418
275 0.5424
300 0.542
325 0.544
350 0.545
375 0.5449
400 0.547
425 0.5456
450 0.547
475 0.5476
};
\addlegendentry{EATA}
\end{axis}

\end{tikzpicture}
        \vspace*{-\baselineskip}
        \caption{Continual TTA methods, under RIP attack.}
        \label{fig:compared_methods_rip_attack}
    \end{subfigure}
    \begin{subfigure}[b]{0.33\textwidth} %
        \centering
        \begin{tikzpicture}
\definecolor{darkgray176}{RGB}{176,176,176}
\definecolor{darkorange}{RGB}{255,140,0}
\definecolor{green01270}{RGB}{0,127,0}
\definecolor{lightgray204}{RGB}{204,204,204}
\definecolor{maroon}{RGB}{128,0,0}
\definecolor{olive}{RGB}{128,128,0}
\definecolor{purple}{RGB}{128,0,128}
\tikzstyle{every node}=[font=\footnotesize]
\begin{axis}[
legend cell align={left},
legend style={
  fill=none,
  fill opacity=0.9,
  draw opacity=1,
  text opacity=1,
  at={(-0.02,1.0)},
  anchor=north west,
  draw=none,
  column sep = 0,
  legend columns=3,
  font=\footnotesize
},
xtick distance=100,
ytick distance=0.1,
xmajorgrids,
ymajorgrids,
tick align=outside,
tick pos=left,
x grid style={darkgray176},
xlabel={Test-time adaptation step $(t)$},
xmin=0, xmax=500,
xtick style={color=black},
y grid style={darkgray176},
ylabel={Testing Error},
xtick align=inside,
ytick align=inside,
xticklabel style = {yshift=-1.0pt, font=\footnotesize},
yticklabel style = {xshift=4pt, , font=\footnotesize},
height=6.3cm, width=6.3cm,
ymin=0.3, ymax=1.0,
ytick style={color=black},
outer sep=0pt,
outer xsep=3pt,
every axis x label/.style={
    at={(axis description cs:0.5,-0.18)},
    align = center},
every axis y label/.style={
    rotate=90,
    at={(axis description cs:-0.15,0.5)},
    align = center},
]
\addplot [semithick, blue, mark=square*, mark size=2.5, mark options={solid}]
table {%
0 0.9639
25 0.6733
50 0.608
75 0.5815
100 0.5749
125 0.5315
150 0.5477
175 0.5141
200 0.5136
225 0.4995
250 0.4897
275 0.4844
300 0.4766
325 0.4638
350 0.4797
375 0.4705
400 0.4651
425 0.4658
450 0.4624
475 0.4662
};
\addlegendentry{PeTTA}
\addplot [semithick, red, mark=*, mark size=2.5, mark options={solid}]
table {%
0 0.9935
25 0.7805
50 0.7698
75 0.7915
100 0.7757
125 0.767
150 0.7908
175 0.7821
200 0.7532
225 0.7474
250 0.7202
275 0.7283
300 0.7283
325 0.6944
350 0.711
375 0.6884
400 0.6964
425 0.6927
450 0.6803
475 0.6682
};
\addlegendentry{RoTTA}
\addplot [semithick, black, mark=pentagon*, mark size=2.5, mark options={solid}]
table {%
0 0.4867
25 0.4759
50 0.4696
75 0.4647
100 0.4556
125 0.4538
150 0.4456
175 0.4391
200 0.4317
225 0.4391
250 0.454
275 0.4442
300 0.4326
325 0.4443
350 0.4311
375 0.4341
400 0.4499
425 0.4419
450 0.422
475 0.4339
};
\addlegendentry{ROID}
\addplot [semithick, olive, mark=triangle*, mark size=2.5, mark options={solid}]
table {%
0 0.9871
25 0.7717
50 0.75
75 0.7623
100 0.759
125 0.7351
150 0.7582
175 0.755
200 0.7435
225 0.7587
250 0.7561
275 0.7497
300 0.7468
325 0.7196
350 0.7214
375 0.7306
400 0.7272
425 0.7323
450 0.7092
475 0.7119
};
\addlegendentry{TRIBE}
\addplot [semithick, purple, mark=diamond*, mark size=2.5, mark options={solid}]
table {%
0 0.483
25 0.4614
50 0.452
75 0.4432
100 0.444
125 0.4488
150 0.441
175 0.4442
200 0.444
225 0.4395
250 0.4393
275 0.4452
300 0.435
325 0.4438
350 0.4439
375 0.442
400 0.443
425 0.4445
450 0.4391
475 0.4364
};
\addlegendentry{CoTTA}
\addplot [semithick, darkorange, mark=+, mark size=2.5, mark options={solid}]
table {%
0 0.488
25 0.4472
50 0.4351
75 0.428
100 0.419
125 0.4257
150 0.4229
175 0.4163
200 0.419
225 0.4137
250 0.4123
275 0.4171
300 0.413
325 0.4091
350 0.4139
375 0.416
400 0.408
425 0.4106
450 0.4159
475 0.4096
};
\addlegendentry{RMT}
\addplot [semithick, maroon, mark=x, mark size=2.5, mark options={solid}]
table {%
0 0.487
25 0.4841
50 0.4783
75 0.4801
100 0.482
125 0.4823
150 0.4755
175 0.4636
200 0.455
225 0.4503
250 0.4509
275 0.457
300 0.467
325 0.4687
350 0.4631
375 0.4535
400 0.448
425 0.45
450 0.4577
475 0.4633
};
\addlegendentry{EATA}
\end{axis}

\end{tikzpicture}
        \vspace*{-\baselineskip}
        \caption{Continual TTA methods, no attack.}
        \label{fig:compared_methods_no_attack}
    \end{subfigure}
    \begin{subfigure}[b]{0.33\textwidth} %
        \centering
        \begin{tikzpicture}
\definecolor{darkgray176}{RGB}{176,176,176}
\definecolor{green01270}{RGB}{0,127,0}
\definecolor{lightgray204}{RGB}{204,204,204}
\definecolor{purple}{RGB}{128,0,128}
\definecolor{teal}{RGB}{0,128,128}
\tikzstyle{every node}=[font=\footnotesize]
\begin{axis}[
legend cell align={left},
legend style={
  fill=white,
  fill opacity=0.9,
  draw opacity=1,
  text opacity=1,
  at={(0.1,0.28)},
  anchor=north west,
  draw=none,
  column sep = 3,
  legend columns=2,
  font=\footnotesize
},
height=6.3cm, width = 6.3 cm,
xtick distance=100,
ytick distance=0.1,
tick align=outside,
xmajorgrids,
ymajorgrids,
tick pos=left,
x grid style={darkgray176},
xlabel={Test-time adaptation step $(t)$},
xmin=0, xmax=501,
xtick style={color=black},
y grid style={darkgray176},
ylabel={Testing Error},
xtick align=inside,
ytick align=inside,
xticklabel style = {yshift=-1.0pt, font=\footnotesize},
yticklabel style = {xshift=4pt, , font=\footnotesize},
ymin=0.3, ymax=0.8,
ytick style={color=black},
outer sep=0pt,
outer xsep=3pt,
every axis x label/.style={
    at={(axis description cs:0.5,-0.18)},
    align = center},
every axis y label/.style={
    rotate=90,
    at={(axis description cs:-0.15,0.5)},
    align = center},
]
\addplot [semithick, red, mark=*, mark size=2.5, mark options={solid}]
table {%
0 0.4865
25 0.4929
50 0.5129
75 0.5174
100 0.5129
125 0.5108
150 0.5107
175 0.5105
200 0.5105
225 0.5118
250 0.5126
275 0.5133
300 0.5151
325 0.516
350 0.5168
375 0.5175
400 0.518
425 0.5183
450 0.5191
475 0.5194
};
\addlegendentry{$\mathcal{L}_{\mathrm{Ent}}$}
\addplot [semithick, blue, mark=square*, mark size=2.5, mark options={solid}]
table {%
0 0.485
25 0.4921
50 0.524
75 0.5684
100 0.617
125 0.6479
150 0.6683
175 0.6807
200 0.688
225 0.6816
250 0.6874
275 0.6937
300 0.695
325 0.6981
350 0.6997
375 0.7012
400 0.703
425 0.7011
450 0.7073
475 0.7093
};
\addlegendentry{$\mathcal{L}_{\mathrm{CE}}$}
\addplot [semithick, purple, mark=diamond*, mark size=2.5, mark options={solid}]
table {%
0 0.4874
25 0.4964
50 0.5269
75 0.5581
100 0.5724
125 0.5829
150 0.5951
175 0.6041
200 0.6092
225 0.6142
250 0.6153
275 0.6232
300 0.6299
325 0.6262
350 0.6325
375 0.6394
400 0.6408
425 0.6436
450 0.6459
475 0.6474
};
\addlegendentry{$\mathcal{L}_{\mathrm{RMT}}$}
\addplot [semithick, teal, mark=pentagon*, mark size=2.5, mark options={solid}]
table {%
0 0.485
25 0.5002
50 0.5149
75 0.5137
100 0.5095
125 0.5044
150 0.5001
175 0.4993
200 0.4989
225 0.4994
250 0.5
275 0.502
300 0.5021
325 0.5022
350 0.5026
375 0.5025
400 0.5028
425 0.5031
450 0.5033
475 0.5042
};
\addlegendentry{$\mathcal{L}_{\mathrm{SLR}}$}
\end{axis}

\end{tikzpicture}
        \vspace*{-\baselineskip}
        \caption{Varying the loss functions.}
        \label{fig:effect_choices_of_loss}
    \end{subfigure}%

    \begin{subfigure}[b]{0.33\textwidth} %
        \centering
        \begin{tikzpicture}
\definecolor{coral}{RGB}{255,127,80}
\definecolor{darkgray176}{RGB}{176,176,176}
\definecolor{darkred}{RGB}{139,0,0}
\definecolor{green01270}{RGB}{0,127,0}
\definecolor{lightgray204}{RGB}{204,204,204}
\definecolor{orangered}{RGB}{255,69,0}
\definecolor{tomato}{RGB}{255,99,71}
\tikzstyle{every node}=[font=\footnotesize]
\begin{axis}[
legend cell align={left},
legend style={
  fill=white,
  fill opacity=0.9,
  draw opacity=1,
  text opacity=1,
  at={(0.02,0.35)},
  anchor=north west,
  draw=none,
  column sep = 3,
  legend columns=2,
  font=\footnotesize
},
height=6.3cm, width=6.3cm,
xtick distance=100,
ytick distance=0.1,
xmajorgrids,
ymajorgrids,
tick align=outside,
tick pos=left,
x grid style={darkgray176},
xlabel={Test-time adaptation step $(t)$},
xmin=0, xmax=501,
xtick style={color=black},
y grid style={darkgray176},
ylabel={Testing Error},
xtick align=inside,
ytick align=inside,
xticklabel style = {yshift=-1.0pt, font=\footnotesize},
yticklabel style = {xshift=4pt, , font=\footnotesize},
ymin=0.3, ymax=0.8,
ytick style={color=black},
outer sep=0pt,
outer xsep=3pt,
every axis x label/.style={
    at={(axis description cs:0.5,-0.18)},
    align = center},
every axis y label/.style={
    rotate=90,
    at={(axis description cs:-0.15,0.5)},
    align = center},
]
]
\addplot [semithick, tomato, mark=triangle*, mark size=2.0, mark options={solid}]
table {%
0 0.49
25 0.495
50 0.5277
75 0.5472
100 0.548
125 0.5505
150 0.5534
175 0.5519
200 0.552
225 0.5557
250 0.5559
275 0.5611
300 0.561
325 0.5609
350 0.5624
375 0.5628
400 0.559
425 0.5608
450 0.5619
475 0.5639
};
\addlegendentry{Level 1}
\addplot [semithick, coral, mark=+, mark size=2.0, mark options={solid}]
table {%
0 0.488
25 0.4951
50 0.5337
75 0.5538
100 0.559
125 0.5669
150 0.5698
175 0.5725
200 0.575
225 0.5748
250 0.5787
275 0.5795
300 0.584
325 0.5819
350 0.5825
375 0.5827
400 0.581
425 0.5879
450 0.588
475 0.5817
};
\addlegendentry{Level 2}
\addplot [semithick, orangered, mark=Mercedes star*, mark size=2.0, mark options={solid,rotate=90}]
table {%
0 0.487
25 0.5031
50 0.5435
75 0.5725
100 0.585
125 0.5835
150 0.5857
175 0.5856
200 0.583
225 0.5869
250 0.5883
275 0.5803
300 0.586
325 0.5862
350 0.5861
375 0.5908
400 0.592
425 0.5912
450 0.5865
475 0.5854
};
\addlegendentry{Level 3}
\addplot [semithick, red, mark=square*, mark size=2.0, mark options={solid}]
table {%
0 0.484
25 0.4923
50 0.535
75 0.5823
100 0.613
125 0.6371
150 0.6495
175 0.6576
200 0.66
225 0.6629
250 0.6626
275 0.6681
300 0.664
325 0.6643
350 0.6661
375 0.6682
400 0.67
425 0.6633
450 0.669
475 0.6715
};
\addlegendentry{Level 4}
\addplot [semithick, darkred, mark=*, mark size=2.0, mark options={solid}]
table {%
0 0.485
25 0.4921
50 0.524
75 0.5684
100 0.617
125 0.6479
150 0.6683
175 0.6807
200 0.688
225 0.6816
250 0.6874
275 0.6937
300 0.695
325 0.6981
350 0.6997
375 0.7012
400 0.703
425 0.7011
450 0.7073
475 0.7093
};
\addlegendentry{Level 5}
\end{axis}

\end{tikzpicture}
        \vspace*{-\baselineskip}
        \caption{Varying data augmentation levels.}
        \label{fig:effect_data_augmentation}
    \end{subfigure}
    \begin{subfigure}[b]{0.33\textwidth} %
        \centering
        \begin{tikzpicture}
\definecolor{darkgray176}{RGB}{176,176,176}
\definecolor{darkorange}{RGB}{255,140,0}
\definecolor{lightgray204}{RGB}{204,204,204}
\definecolor{maroon}{RGB}{128,0,0}
\tikzstyle{every node}=[font=\footnotesize]
\begin{axis}[
legend cell align={left},
legend style={
  fill=white,
  fill opacity=0.9,
  draw opacity=1,
  text opacity=1,
  at={(0.1,0.2)},
  anchor=north west,
  draw=none,
  column sep = 3,
  legend columns=2,
  font=\footnotesize
},
height=6.3cm, width=6.3cm,
xtick distance=100,
ytick distance=0.1,
tick align=outside,
xmajorgrids,
ymajorgrids,
tick pos=left,
x grid style={darkgray176},
xlabel={Test-time adaptation step $(t)$},
xmin=0, xmax=501,
xtick style={color=black},
y grid style={darkgray176},
ylabel={Testing Error},
xtick align=inside,
ytick align=inside,
xticklabel style = {yshift=-1.0pt, font=\footnotesize},
yticklabel style = {xshift=4pt, , font=\footnotesize},
ymin=0.3, ymax=1.0,
ytick style={color=black},
outer sep=0pt,
outer xsep=3pt,
every axis x label/.style={
    at={(axis description cs:0.5,-0.18)},
    align = center},
every axis y label/.style={
    rotate=90,
    at={(axis description cs:-0.15,0.5)},
    align = center},
]
]
\addplot [semithick, maroon, mark=x, mark size=2.5, mark options={solid}]
table {%
0 0.485
25 0.4815
50 0.4996
75 0.5374
100 0.573
125 0.6305
150 0.6784
175 0.7317
200 0.777
225 0.8105
250 0.8535
275 0.8886
300 0.915
325 0.9367
350 0.9514
375 0.965
400 0.97
425 0.9748
450 0.9786
475 0.9802
};
\addlegendentry{Student}
\addplot [semithick, darkorange, mark=+, mark size=2.5, mark options={solid}]
table {%
0 0.485
25 0.4784
50 0.4749
75 0.4833
100 0.483
125 0.4908
150 0.4851
175 0.484
200 0.488
225 0.4939
250 0.4995
275 0.5014
300 0.506
325 0.5154
350 0.5247
375 0.5267
400 0.537
425 0.5375
450 0.5453
475 0.5502
};
\addlegendentry{Teacher}
\end{axis}

\end{tikzpicture}
        \vspace*{-\baselineskip}
        \caption{Two choices of pseudo-label predictor.}
        \label{fig:effect_choices_of_pseudo_lbl_predictor}
    \end{subfigure}%
    \begin{subfigure}[b]{0.33\textwidth}
        \centering
        \begin{tikzpicture}
\definecolor{darkgray176}{RGB}{176,176,176}
\definecolor{darkorange}{RGB}{255,140,0}
\definecolor{lightgray204}{RGB}{204,204,204}
\definecolor{maroon}{RGB}{128,0,0}
\definecolor{purple}{RGB}{128,0,128}
\definecolor{teal}{RGB}{0,128,128}
\tikzstyle{every node}=[font=\footnotesize]
\begin{axis}[
legend cell align={left},
legend style={
  fill=white,
  fill opacity=0.9,
  draw opacity=1,
  text opacity=1,
  at={(0.02,0.3)},
  anchor=north west,
  draw=none,
  column sep = 3,
  legend columns=2,
  font=\footnotesize
},
height=6.3cm, width=6.3cm,
xtick distance=100,
ytick distance=0.1,
tick align=outside,
xmajorgrids,
ymajorgrids,
tick pos=left,
x grid style={darkgray176},
xlabel={Test-time adaptation step $(t)$},
xmin=0, xmax=501,
xtick style={color=black},
y grid style={darkgray176},
ylabel={Testing Error},
xtick align=inside,
ytick align=inside,
xticklabel style = {yshift=-1.0pt, font=\footnotesize},
yticklabel style = {xshift=4pt, , font=\footnotesize},
ymin=0.2, ymax=1.0,
ytick style={color=black},
outer sep=0pt,
outer xsep=3pt,
every axis x label/.style={
    at={(axis description cs:0.5,-0.18)},
    align = center},
every axis y label/.style={
    rotate=90,
    at={(axis description cs:-0.15,0.5)},
    align = center},
]
]
]
\addplot [semithick, maroon, mark=x, mark size=2.5, mark options={solid}]
table {%
0 0.485
25 0.4811
50 0.4989
75 0.5376
100 0.569
125 0.6257
150 0.6755
175 0.7296
200 0.771
225 0.808
250 0.8469
275 0.8827
300 0.905
325 0.9279
350 0.9452
375 0.9558
400 0.962
425 0.9678
450 0.9724
475 0.9742
};
\addlegendentry{$\alpha=0.0$}
\addplot [semithick, darkorange, mark=+, mark size=2.5, mark options={solid}]
table {%
0 0.485
25 0.4808
50 0.4965
75 0.5315
100 0.558
125 0.6068
150 0.6589
175 0.7066
200 0.741
225 0.7807
250 0.8302
275 0.8633
300 0.892
325 0.9164
350 0.937
375 0.9542
400 0.965
425 0.9717
450 0.9787
475 0.9819
};
\addlegendentry{$\alpha=0.5$}
\addplot [semithick, red, mark=*, mark size=2.5, mark options={solid}]
table {%
0 0.485
25 0.4792
50 0.4829
75 0.5061
100 0.522
125 0.5452
150 0.5712
175 0.6003
200 0.621
225 0.6419
250 0.6773
275 0.6976
300 0.719
325 0.7445
350 0.7843
375 0.8088
400 0.838
425 0.8599
450 0.881
475 0.8933
};
\addlegendentry{$\alpha=0.9$}
\addplot [semithick, blue, mark=square*, mark size=2.5, mark options={solid}]
table {%
0 0.485
25 0.4786
50 0.4804
75 0.5001
100 0.506
125 0.52
150 0.5357
175 0.5527
200 0.565
225 0.5828
250 0.5941
275 0.6116
300 0.622
325 0.6395
350 0.6569
375 0.6822
400 0.706
425 0.7269
450 0.742
475 0.7597
};
\addlegendentry{$\alpha=0.95$}
\addplot [semithick, purple, mark=diamond*, mark size=2.5, mark options={solid}]
table {%
0 0.485
25 0.4785
50 0.475
75 0.4824
100 0.482
125 0.4904
150 0.4849
175 0.4833
200 0.488
225 0.4929
250 0.4946
275 0.4988
300 0.504
325 0.5146
350 0.5218
375 0.523
400 0.532
425 0.5293
450 0.5367
475 0.5422
};
\addlegendentry{$\alpha=0.99$}
\addplot [semithick, teal, mark=pentagon*, mark size=2.5, mark options={solid}]
table {%
0 0.485
25 0.4781
50 0.4741
75 0.4787
100 0.476
125 0.479
150 0.474
175 0.4712
200 0.479
225 0.4761
250 0.4796
275 0.4758
300 0.477
325 0.4744
350 0.4779
375 0.4737
400 0.479
425 0.4771
450 0.4718
475 0.4734
};
\addlegendentry{$\alpha=1.0$}
\end{axis}

\end{tikzpicture}
        \vspace*{-\baselineskip}
        \caption{EMA with varying update rates ($\alpha$, \Eq{\ref{eq:general_teacher_update}}).}
        \label{fig:effect_update_rate}
    \end{subfigure}
    \caption{Average of the \metric in various CIFAR-10-C~\cite{hendrycks2019robustness} experiments. 
    (a) The testing error of several major TTA algorithms~\cite{yuan2023robust, döbler2023robust, su2023realworld, marsden2024universal} progressively raises under RIP attack while some early-generation algorithms~\cite{niu2022efficient, Wang_2022_CVPR} persist, surprisingly.
    (b) This was not the case with no attack (normal condition).
    Ablation studies on a baseline model, trained with \Lss{CE} (with augmentation) and update rate $\alpha=0.99$ are conducted. 
    The plot (c) confirms the risk of RIP attack on TTA methods using data augmentation in their loss function, (d) the stronger the augmentation, the more vulnerable. 
    (e) Using mean-teacher (EMA) model update, or (f) slowing down the update rate can mitigate the effect (compared to $\alpha=0$, no EMA update), but fail to eliminate it (considering $\alpha=1$, no TTA). 
    }
    \label{fig:results_rip_causes_of_vulnerability}
\end{figure*}

\noindent
The vulnerability of many existing TTA methods under RIP attack is confirmed in \Tab{\ref{tab:rip_attack_compared_methods}}. 
\textit{Overall, the average testing error increment is observed in all datasets}.
To qualitatively observe this effect on CIFAR-10-C~\cite{hendrycks2019robustness}, we visualize the testing error after every $25$ steps. Two scenarios are considered: under RIP attack (\Fig{\ref{fig:compared_methods_rip_attack}}), and no attack (\Fig{\ref{fig:compared_methods_no_attack}}). 
Surprisingly, CoTTA~\cite{Wang_2022_CVPR} or EATA~\cite{niu2022efficient} - the earliest methods have the best resilience to RIP.
The following \Sec{\ref{sec:causes_of_risks}} conducts ablation studies to explain when a TTA method fails or thrives.
As a teaser for their resilience, EATA does not involve training with augmented sample \texttt{Aug}, CoTTA is a simple method that uses the teacher model for predicting the pseudo-labels (\Eq{\ref{eq:pseudo_from_teacher}}). 
However, we note that while their limitations and assumptions do exist, motivating the development of many subsequent methods~\cite{gong2022note, yuan2023robust}.
\textit{The behavior of real TTA methods matches the risk} (\Sec{\ref{sec:theoretical}}).
Additional attack results are provided in the Appendix.

\section{Analyzing the Causes of Vulnerability}
\label{sec:causes_of_risks}

There are multiple living components inside a continual TTA. This section introduces a baseline method (\Sec{\ref{sec:baseline_method}}) and a series of ablation studies, isolating those factors to evaluate the risk of existing design choices: loss function (\Sec{\ref{sec:study_loss_function}}), level of augmentation (\Sec{\ref{sec:study_level_augmentation}}), pseudo-label generator (\Sec{\ref{sec:study_pseudo_label}}) and model update scheme (\Sec{\ref{sec:study_update_schemes}}).

\subsection{Baseline Continual TTA Method}
\label{sec:baseline_method}
We employ a baseline continual TTA method based on \Eq{\ref{eq:aug_pseudo_lbl}, \ref{eq:general_teacher_update}}, closed to CoTTA~\cite{Wang_2022_CVPR}. 
This simple model updates the linear parameters of batch normalization~\cite{pmlr-v37-ioffe15, wang2021tent}. 
For simplicity, the ablation studies are conducted on CIFAR-10-C~\cite{hendrycks2019robustness} - $10$ trials averaged.
Without specifically noted, the $\mathcal{L}_{\mathrm{CE}}$ with augmentation and update rate $\alpha=0.99$ are used.

\subsection{Effects of the Loss Function Choices }
\label{sec:study_loss_function}

\noindent \textbf{Setup}. We explore the effect of the loss function on the robustness of a TTA method under RIP attack. 
Different choices of loss functions in \Sec{\ref{sec:tta_loss_functions}}: 
$\mathcal{L}_{\mathrm{Ent}}$ (\Eq{\ref{eq:ent_loss}}),
$\mathcal{L}_{\mathrm{CE}}$ (\Eq{\ref{eq:ce_loss}}, augmentation),
$\mathcal{L}_{\mathrm{RMT}}$~\cite{döbler2023robust},
and $\mathcal{L}_{\mathrm{SLR}}$ (\Eq{\ref{eq:slr_loss}})
are used as $\mathcal{L}_{\mathrm{CLS}}$ in the baseline method's update step (\Eq{\ref{eq:general_opti_step}}). 

\noindent \textbf{Results.}
\Fig{\ref{fig:effect_choices_of_loss}} introduces the testing errors. 
The effect of RIP on each loss function choice is different. 
Notably,\textit{ all loss functions that involve augmented samples} (\Eq{\ref{eq:aug_pseudo_lbl}}) as discussed in \Sec{\ref{sec:tta_loss_functions}}, such as $\mathcal{L}_{\mathrm{CE}}$ and $\mathcal{L}_{\mathrm{RMT}}$, \textit{are more susceptible to RIP attack}. In contrast, simpler loss functions like $\mathcal{L}_{\mathrm{Ent}}$ and $\mathcal{L}_{\mathrm{SLR}}$ perform well. 
$\mathcal{L}_{\mathrm{RMT}}$, falling between these two categories of functions, is intermediate.

\subsection{Effects of the Level of Augmentation}
\label{sec:study_level_augmentation}

\noindent \textbf{Setup. } To confirm the suspicions of augmented samples, we investigate the correlation between the level of data augmentation and the tolerance of the baseline model to RIP.
Following a prior study~\cite{Wang_2022_CVPR}, random color jitter, affine transformations, and horizontal flipping are applied. The level of augmentation varies from 1 to 5 (5 is the strongest, and the typical level used in practice-\cite{Wang_2022_CVPR}).
Visit the Appendix for visual examples of these augmented samples.

\noindent \textbf{Results. } The effect of the RIP attack on the baseline algorithm is visualized in \Fig{\ref{fig:effect_data_augmentation}}. Unsurprisingly, \textit{the damage of RIP is correlated with the level of augmentation used}, well explained by the shifting boundary effect in \Sec{\ref{sec:ips_gmmc}}.

\subsection{Effects of the Pseudo-label Predictor}
\label{sec:study_pseudo_label}
\textbf{Setup.} Previous ablation studies identified the use of loss functions computed on augmented samples as the main cause of RIP vulnerability. 
\Sec{\ref{sec:pseudo-label-predictor}} presents two choices of model for pseudo-label predictors: teacher (\Eq{\ref{eq:pseudo_from_teacher}}) and student (\Eq{\ref{eq:pseudo_from_student}}). 
They are investigated in this section.  

\noindent \textbf{Results.} The experimental result in \Fig{\ref{fig:effect_choices_of_pseudo_lbl_predictor}} shows that \textit{the pseudo labels predicted by the student model make continual TTA methods more vulnerable to RIP attach, compared to the teacher model}.
This can be explained by the improved accuracy of pseudo-labels for adaptation, as stated in~\cite{antti2017_mean_teachers}.

\subsection{Effects of the Model Update Rate}
\label{sec:study_update_schemes}
\noindent \textbf{Setup. } Extending the observation in \Sec{\ref{sec:study_pseudo_label}}, we study the effect of the update rate $\alpha$ (\Eq{\ref{eq:general_teacher_update}}) on the resilience of the baseline model under RIP attack. Various values are chosen from $\alpha=0$ (no mean teacher update) and $\alpha=1$ (no TTA).

\noindent \textbf{Results.} \Fig{\ref{fig:effect_update_rate}} plots the testing error with increasing value of $\alpha$.
\textit{The slower the update rate, the better the model can mitigate RIP attack}. However, it cannot be eliminated.

\begin{figure}[ht!]
    \centering
    \begin{tikzpicture}

\definecolor{darkgray176}{RGB}{176,176,176}
\definecolor{darkorange}{RGB}{255,140,0}
\definecolor{lightgray204}{RGB}{204,204,204}
\definecolor{maroon}{RGB}{128,0,0}

\tikzstyle{every node}=[font=\footnotesize]
\begin{axis}[
legend cell align={left},
legend style={
  fill=none,
  fill opacity=0.9,
  draw opacity=1,
  text opacity=1,
  at={(0.0,0.4)},
  anchor=north west,
  draw=none,
  column sep = 0,
  legend columns=2,
  font=\footnotesize
},
tick align=outside,
tick pos=left,
x grid style={darkgray176},
xlabel={Test-time adaptation step $(t)$},
xmin=0, xmax=500,
xtick distance=100,
ytick distance=0.1,
xtick style={color=black},
y grid style={darkgray176},
ylabel={Testing Error},
xtick align=inside,
ytick align=inside,
xticklabel style = {yshift=-1.0pt, font=\footnotesize},
yticklabel style = {xshift=4pt, , font=\footnotesize},
ymajorgrids,
xmajorgrids,
height=5.0cm, width = 8.8 cm,
ymin=0.3, ymax=0.75,
ytick style={color=black},
outer sep=0pt,
outer xsep=3pt,
every axis x label/.style={
    at={(axis description cs:0.5,-0.18)},
    align = center},
every axis y label/.style={
    rotate=90,
    at={(axis description cs:-0.1,0.5)},
    align = center},
]
\addplot [semithick, maroon, mark=x, mark size=2.5, mark options={solid}]
table {%
0 0.485
25 0.4921
50 0.524
75 0.5684
100 0.617
125 0.6479
150 0.6683
175 0.6807
200 0.688
225 0.6816
250 0.6874
275 0.6937
300 0.695
325 0.6981
350 0.6997
375 0.7012
400 0.703
425 0.7011
450 0.7073
475 0.7093
};
\addlegendentry{Baseline (B)}
\addplot [semithick, darkorange, mark=+, mark size=2.5, mark options={solid}]
table {%
0 0.483
25 0.4759
50 0.5112
75 0.5447
100 0.575
125 0.5908
150 0.5946
175 0.5909
200 0.591
225 0.5927
250 0.5946
275 0.5982
300 0.59
325 0.5987
350 0.5939
375 0.5945
400 0.6
425 0.5992
450 0.59
475 0.5986
};
\addlegendentry{B + SrcReplay}
\addplot [semithick, red, mark=*, mark size=2.5, mark options={solid}]
table {%
0 0.486
25 0.4829
50 0.5137
75 0.5578
100 0.592
125 0.61
150 0.6175
175 0.6256
200 0.632
225 0.636
250 0.6316
275 0.6387
300 0.64
325 0.6395
350 0.641
375 0.6369
400 0.638
425 0.6395
450 0.6442
475 0.6461
};
\addlegendentry{B + SrcContrast}
\addplot [semithick, blue, mark=square*, mark size=2.5, mark options={solid}]
table {%
0 0.49
25 0.4759
50 0.5171
75 0.5524
100 0.574
125 0.5755
150 0.5741
175 0.5736
200 0.574
225 0.5733
250 0.5782
275 0.5775
300 0.569
325 0.5742
350 0.5757
375 0.5768
400 0.573
425 0.5756
450 0.5722
475 0.5757
};
\addlegendentry{B + SrcContrast + SrcReplay}
\addplot [semithick, purple, mark=diamond*, mark size=2.5, mark options={solid}]
table {%
0 0.486
25 0.4804
50 0.4886
75 0.5111
100 0.518
125 0.5256
150 0.5262
175 0.525
200 0.526
225 0.5327
250 0.5196
275 0.5223
300 0.525
325 0.5264
350 0.5138
375 0.5196
400 0.521
425 0.5186
450 0.5253
475 0.517
};
\addlegendentry{B + SrcEnsemble}
\end{axis}

\end{tikzpicture}
    \vspace*{-1.5\baselineskip}
    \caption{Average testing error of the baseline model (\Sec{\ref{sec:baseline_method}}) with three simple RIP attack defensive attempts: source training replay (SrcReplay), contrastive loss with source prototypes (SrcContrast), and source model weight ensemble (SrcEnsemble).}
    \label{fig:results_defending_rip_attack}
    \vspace*{-\baselineskip}
\end{figure}
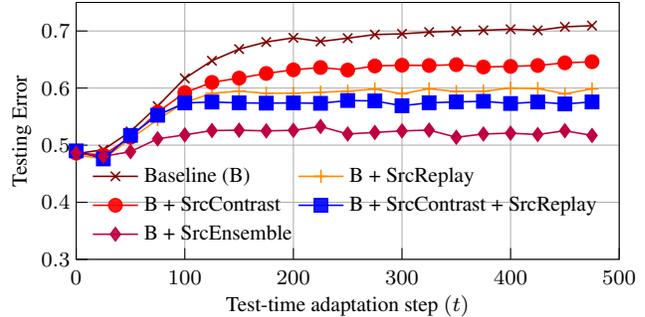

\section{Discussions and Conclusions}
\label{sec:rip_defense}

\noindent \textbf{RIP Attack Defense.} Although this is not the primary focus of this study,
we still investigate some techniques: the source replay~\cite{longji1992, döbler2023robust}, the contrastive loss~\cite{kang2019contrastive, döbler2023robust}, and the source model weight ensemble (\Eq{\ref{eq:source_weights_ensemble_update}}).
\Fig{\ref{fig:results_defending_rip_attack}} visualizes their effect. 
Constraining the model’s divergence from the source model appears to \textit{help preserve the resistance} of the baseline TTA method, but it \textit{does not fully eliminate the risk}.
Eliminating \texttt{Aug} operator could remove the risk, but it comes at the cost of reduced performance.
While using the teacher model for pseudo-label prediction can make the attack more difficult, there is a trade-off as noted in~\cite{park2024medbn}. 

\noindent \textbf{Limitations of RIP Attack.} 
Though it may not be costly to execute, RIP attack still requires either collecting a small labeled dataset or a manual step to find incorrect predictions.
From \Sec{\ref{sec:results}, \ref{sec:causes_of_risks}}, RIP attack can only succeed if a random augmentation operator is used. 
Although this is not a criterion in prior studies~\cite{wu2023uncovering, cong2024poisoning}, we recognize that these attempts, including RIP, require continuous sample submissions, occupying the testing stream for a prolonged period.

\noindent \textbf{Future Work. } The introduced RIP is relatively simple, only utilizing the final predicted label as the feedback signal for the attack purpose. 
An elaborated version of it can be developed by exploiting the output probability or the confidence score associated with each prediction. 
The ability to defend RIP attack is only briefly discussed, necessitating the development of an efficient RIP defense mechanism. 

\noindent \textbf{Conclusions.} Orthogonal to prior TTA attack studies, \textit{Reusing of Incorrect Prediction (RIP)} draws our attention to an untouched concern: \textit{``Continual TTA is vulnerable to an intriguingly simple black-box algorithm"}. 
This study confirms the risk on recent continual TTA methods and highlights that the use of augmentation is correlated to RIP's vulnerability. 
As the key mechanism, TTA makes a model more confident in their predictions after each adaptation step, which is undesirably, also magnifies the errors caused by incorrect pseudo-labels. 
This becomes a \textit{backdoor} for attackers to intentionally collapse a continual TTA model.%

\clearpage
{
    \small
    \bibliographystyle{ieeenat_fullname}
    \bibliography{egbib}
}

\clearpage

\appendix
\normalsize
\begin{center}
\large{\textbf{\paperTitle \\ Supplementary Material}}    
\end{center}

\section{Related Work}
\noindent \textbf{Continual Test-time Adaptation (TTA).} 
Under the circumstance of testing data distribution diverged~\cite{10.5555/1462129}, \textit{Test-Time Adaptation (TTA)}, a domain generalization technique~\cite{kaiyang2023}, enhances the performance of a machine learning (ML) model by enabling its parameters to change through test-time training~\cite{pmlr-v119-sun20b, yuejiang2021_ttt+}. 
Fundamentally, TTA encourages the ML model to be more confident in their predictions by minimizing the prediction entropy~\cite{wang2021tent, niu2022efficient, nguyen2023tipi, wang2021tent, niu2023towards, liang2020we}. 
Observing that the distribution may not happen once, but can be changed continuously, later studies extend the TTA approach to multiple shifts setting~\cite{Wang_2022_CVPR, press2023rdumb, hoang2024petta}.
Towards real-world TTA, recent research studies TTA also investigate the scenarios where the label distribution is non-i.i.d., or temporally correlated ~\cite{gong2022note, yuan2023robust, niu2023towards, marsden2024universal}. 
These studies, on the one hand, address major failure modes of TTA on challenging testing streams, are far more complicated than earlier methods such as~\cite{Wang_2022_CVPR, niu2022efficient}. 
This necessitates further investigation into their reliability and trustworthiness during deployment, \textit{with RIP attacks being on this line of inquiry}.

\noindent \textbf{Adversarial Attacks and Defenses in ML. } 
An adversarial attack aims to degrade the performance of a victim ML model by manipulating the input data, causing it to produce false predictions~\cite{yulong2023, szegedy2014, goodfellow2015, nguyen2015}. 
Those attempts can be classified into main categories: white-box, black-box, and gray-box~\cite{xu2021-greybox}, based on the attacker’s knowledge of the victim system.  
The white-box attack assumes the model is fully accessible, allowing gradient-based algorithms to craft the adversarial samples~\cite{szegedy2014, goodfellow2015}.
Meanwhile, the black-box attack strictly restricts these favors and repetitively queries the model to adjust the strategy~\cite{nicolas2017_practical}.  
\textit{A black-box attack is significantly more realistic.}
Adversarial defense techniques are also explored to counteract attack attempts.
Straightforwardly, a trained ML model to detect and filter adversarial samples~\cite{liang2021detecting, tianyu2018, yulong2023}.
More advanced techniques can make the model more robust to such perturbations during training~\cite{papernot2016, sven2021} or be more adaptive at test time~\cite{Croce2022EvaluatingTA}.

\noindent \textbf{Continual TTA Attack.} 
While adversarial attack and continual TTA are two active research areas, only a limited number of prior studies investigate this risk on continual TTA methods.
Poisoning~\cite{cong2024poisoning} or distribution invading~\cite{wu2023uncovering, park2024medbn} attacks are among the first studies uncovering the risk of continual TTA being manipulated by injecting malicious testing samples. 
Even though they are the closest to our work, both fall into the \textit{white-box attack category} that assumes the attacker has complete knowledge of the underlying model. 
In contrast, our RIP attack neither makes any assumptions about the operating TTA algorithm nor the adaptation process, and we simply audit the model outputs to modify the attack scheme.
To the best of our knowledge, \textit{we are the first to introduce an implementable black-box attack algorithm targeting a continual TTA method}. 

\noindent \textbf{Comparison of RIP Versus Previous Continual TTA Attack Studies. } Besides studying a black-box continual TTA attack for the first time, \textit{this work is orthogonal to the prior studies}~\cite{cong2024poisoning, wu2023uncovering, park2024medbn} \textit{in multiple aspects}. 
Firstly, the investigation here primarily focuses on the risk during self-training (data augmentation, loss function, model update, and pseudo-label predictor) utilized for the TTA process, not the risk in intercepting the batch normalization~\cite{pmlr-v37-ioffe15} statistics like~\cite{wu2023uncovering, park2024medbn} to create false predictions. 
The long-term effect on a TTA model after the attack is not also well investigated.
Secondly, this work targets the recent \textit{continual} TTA algorithms, mean-teacher model update~\cite{antti2017_mean_teachers}, not the traditional single-domain TTA approaches~\cite{wang2021tent, niu2023towards, nguyen2023tipi}.
Thirdly, the attack action of RIP is inherently natural, as it simply reuses images from the testing distribution without requiring any image-level modifications via adversarial training and gradient descent update~\cite{szegedy2014, goodfellow2015}.

\section{Additional RIP Attack Results}
In \Alg{\ref{alg:rip_attack}}, all the input parameters of RIP attack have been introduced. 
Within the main text, the effectiveness of RIP has been demonstrated on a variety of different continual TTA methods (varying $f_t(x)$) (\Sec{\ref{sec:results_existingTTA}}). 
While the number of attack steps $T_a$ can be monitored by determining whether the model performance is saturated, this section provides additional RIP attack results to complete the discussion. Specifically, we study RIP attack with varying victim class $y_a$ (\Sec{\ref{sec:different_victim_class}}), labeled attack dataset $\mathcal{D}_a$ (\Sec{\ref{sec:different_attack_datasets}}) and testing batch size $B$ (\Sec{\ref{sec:different_batch_sizes}}).

\subsection{Varying Victim Classes}
\label{sec:different_victim_class}
\begin{figure*}[ht!]
    \centering
    \begin{subfigure}[t]{.97\linewidth}
        \centering
        \includegraphics[width=\linewidth]{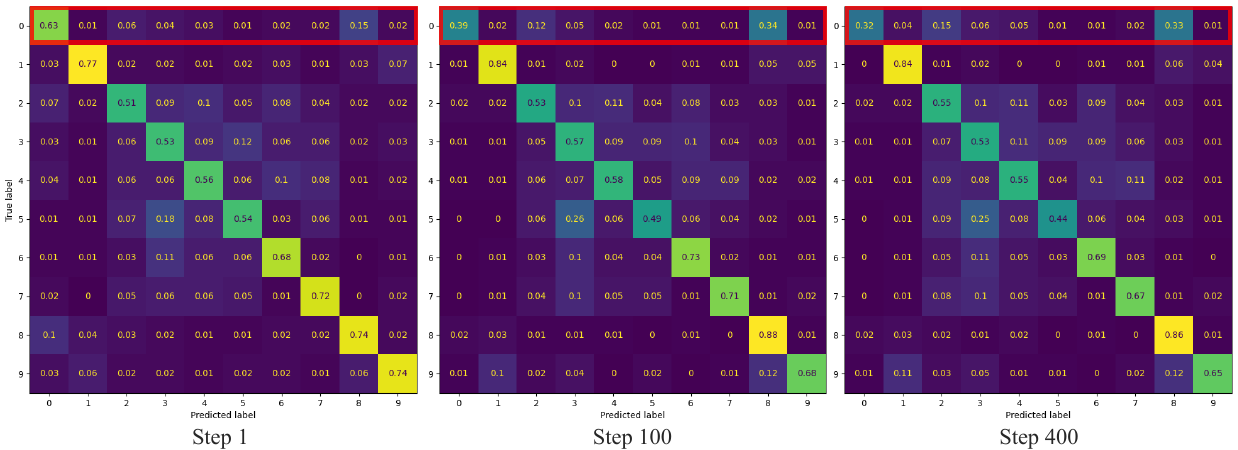}
        \caption{Victim class $y_a=0$ \textit{(airplane)}}
    \end{subfigure}%
    
    \begin{subfigure}[t]{.97\linewidth}
        \centering
        \includegraphics[width=\linewidth]{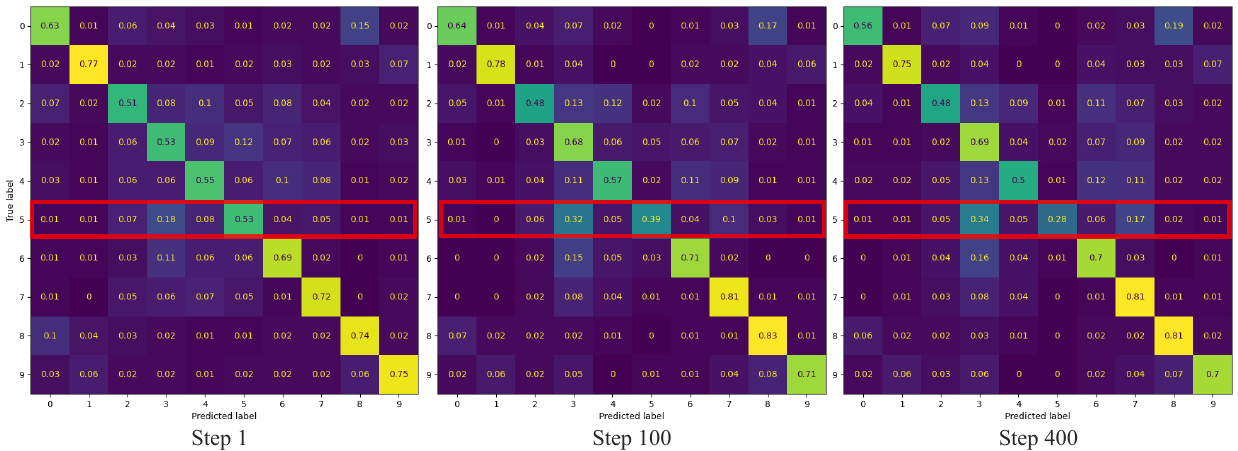}
        \caption{Victim class $y_a=5$ \textit{(dog)}}
    \end{subfigure}

     \begin{subfigure}[t]{.97\linewidth}
        \centering
        \includegraphics[width=\linewidth]{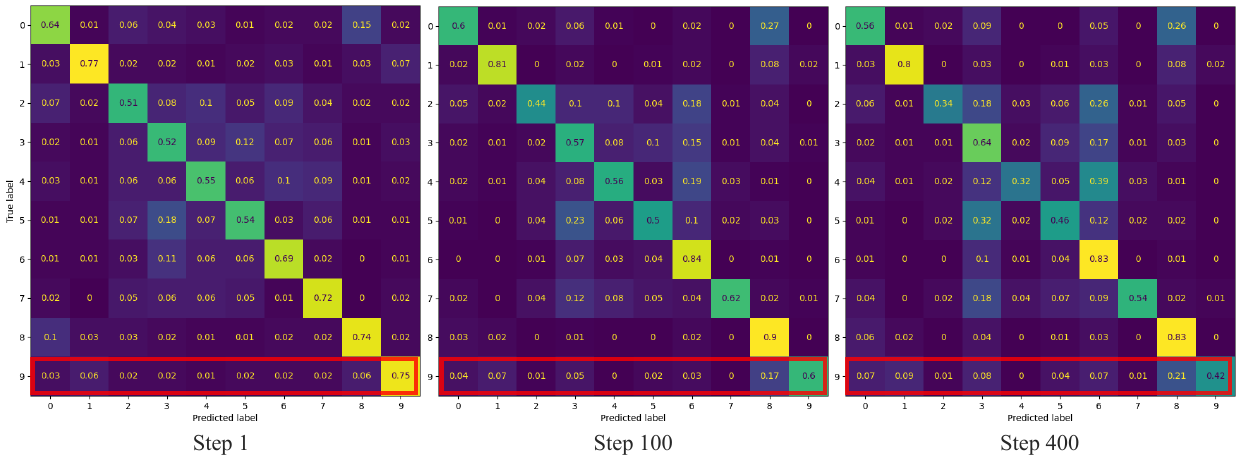}
        \caption{Victim class $y_a=9$ \textit{(truck)}}
    \end{subfigure}
    \caption{Confusion matrices under RIP attack of the baseline continual TTA method (\Sec{\ref{sec:baseline_method}}) at the $1^{st}$, $100^{th}$ and the $400^{th}$ step. 
    CIFAR-10-C~\cite{hendrycks2021} is used with three choices of victim class: (a) $y_a=0$ \textit{(airplane)}, (b) $y_a=5$ \textit{(dog)}, and (c) $y_a=9$ \textit{(truck)} for demonstration, the other classes follow similarly.
    The rows presenting the true labels are highlighted in \textcolor{red}{red} boxes.
    We direct readers' attention to the victim class and the entry in this row most susceptible to misclassification. 
    As adaptation proceeds, the model increasingly tends to \textit{further misclassify this entry as the victim class}. 
    This observation agrees with the \textit{shifting boundary effect} (\Fig{\ref{fig:rip-cartoon}} and \Sec{\ref{sec:ips_gmmc}}).
    }
    \label{fig:cfs_mat_rip_attack}
\end{figure*}

We provide the confusion matrices of the baseline continual TTA method (\Sec{\ref{sec:baseline_method}}) under RIP attack in \Fig{\ref{fig:cfs_mat_rip_attack}}.
Here, the confusion matrix of this model at the $1^{st}$, $100^{th}$, and the $400^{th}$ adaptation step is visualized.
From this figure, we can see the effect of RIP attack with \textit{three different choices of the victim class} $y_a$. 
It is hard to see in the main tables and figures since they provide the averaged value across all $10$ selected classes (or trials, as defined in the main text).
Nevertheless, the observation is similar, regardless of the selected victim class.
Furthermore, this figure also showcases the \textit{``shifting boundary effect"} (\Fig{\ref{fig:rip-cartoon}} and \Sec{\ref{sec:ips_gmmc}}) on a real image dataset.
A transition in the model predictions on the victim class to the most misclassified class is consistently observed in all the choices of the victim class.

\subsection{Varying Choices of Labeled Attack Dataset}
\label{sec:different_attack_datasets}

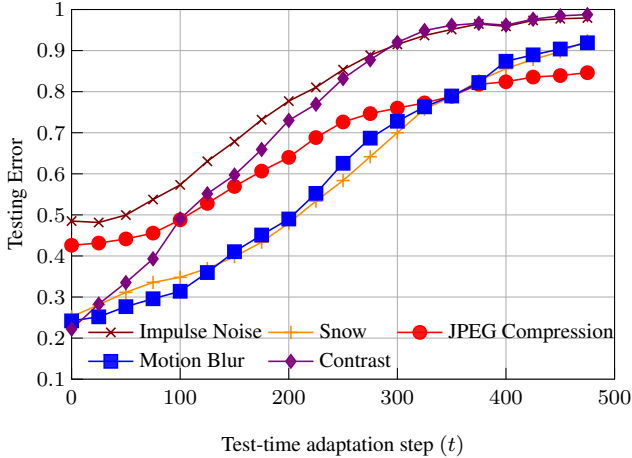
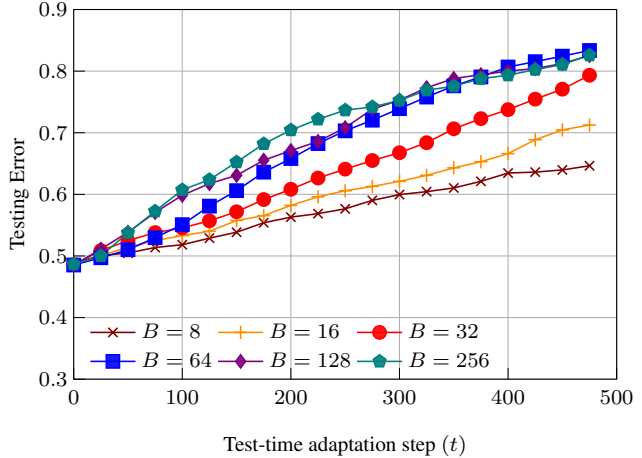
\begin{figure*}[ht!]
    \begin{subfigure}[t]{.48\linewidth}
        \centering
        \begin{tikzpicture}

\definecolor{darkgray176}{RGB}{176,176,176}
\definecolor{darkorange}{RGB}{255,140,0}
\definecolor{lightgray204}{RGB}{204,204,204}
\definecolor{maroon}{RGB}{128,0,0}
\definecolor{purple}{RGB}{128,0,128}

\tikzstyle{every node}=[font=\footnotesize]
\begin{axis}[
legend cell align={left},
legend style={
  fill=none,
  fill opacity=0.9,
  draw opacity=1,
  text opacity=1,
  at={(0.0,0.18)},
  anchor=north west,
  draw=none,
  column sep = 0,
  legend columns=3,
  font=\footnotesize
},
tick align=outside,
tick pos=left,
x grid style={darkgray176},
xlabel={Test-time adaptation step $(t)$},
xmin=0, xmax=500,
xtick distance=100,
ytick distance=0.1,
xtick style={color=black},
y grid style={darkgray176},
ylabel={Testing Error},
xtick align=inside,
ytick align=inside,
xticklabel style = {yshift=-1.0pt, font=\footnotesize},
yticklabel style = {xshift=4pt, , font=\footnotesize},
ymajorgrids,
xmajorgrids,
height=6.5cm, width = 8.8 cm,
ymin=0.1, ymax=1.0,
ytick style={color=black},
outer sep=0pt,
outer xsep=3pt,
every axis x label/.style={
    at={(axis description cs:0.5,-0.18)},
    align = center},
every axis y label/.style={
    rotate=90,
    at={(axis description cs:-0.1,0.5)},
    align = center},
]
\addplot [semithick, maroon, mark=x, mark size=2.5, mark options={solid}]
table {%
0 0.485
25 0.4815
50 0.4996
75 0.5374
100 0.573
125 0.6305
150 0.6784
175 0.7317
200 0.777
225 0.8105
250 0.8535
275 0.8886
300 0.915
325 0.9367
350 0.9514
375 0.9649
400 0.959
425 0.9735
450 0.9775
475 0.9794
};
\addlegendentry{Impulse Noise}
\addplot [semithick, darkorange, mark=+, mark size=2.5, mark options={solid}]
table {%
0 0.252
25 0.2812
50 0.3114
75 0.3356
100 0.348
125 0.3694
150 0.3982
175 0.4336
200 0.48
225 0.5334
250 0.5836
275 0.6416
300 0.7
325 0.7574
350 0.7876
375 0.8266
400 0.856
425 0.8786
450 0.8974
475 0.924
};
\addlegendentry{Snow}
\addplot [semithick, red, mark=*, mark size=2.5, mark options={solid}]
table {%
0 0.426
25 0.4314
50 0.4414
75 0.4556
100 0.488
125 0.528
150 0.5692
175 0.6066
200 0.64
225 0.6882
250 0.7262
275 0.7466
300 0.76
325 0.7726
350 0.7888
375 0.8178
400 0.824
425 0.8356
450 0.8392
475 0.8462
};
\addlegendentry{JPEG Compression}
\addplot [semithick, blue, mark=square*, mark size=2.5, mark options={solid}]
table {%
0 0.242
25 0.2518
50 0.2766
75 0.2958
100 0.314
125 0.3596
150 0.4104
175 0.451
200 0.49
225 0.5524
250 0.6256
275 0.6868
300 0.728
325 0.763
350 0.7894
375 0.8222
400 0.874
425 0.8896
450 0.9038
475 0.919
};
\addlegendentry{Motion Blur}
\addplot [semithick, purple, mark=diamond*, mark size=2.5, mark options={solid}]
table {%
0 0.222
25 0.2828
50 0.3352
75 0.393
100 0.49
125 0.5514
150 0.597
175 0.6592
200 0.73
225 0.7694
250 0.8318
275 0.8776
300 0.92
325 0.9488
350 0.9616
375 0.9664
400 0.962
425 0.9764
450 0.9846
475 0.9876
};
\addlegendentry{Contrast}
\end{axis}

\end{tikzpicture}
        \caption{Various choices of the attack dataset $\mathcal{D}_a$.}
        \label{fig:results_corruptions_for_attack}
    \end{subfigure}
    \begin{subfigure}[t]{.48\linewidth}
        \centering
        \begin{tikzpicture}

\definecolor{darkgray176}{RGB}{176,176,176}
\definecolor{darkorange}{RGB}{255,140,0}
\definecolor{lightgray204}{RGB}{204,204,204}
\definecolor{maroon}{RGB}{128,0,0}
\definecolor{purple}{RGB}{128,0,128}
\definecolor{teal}{RGB}{0,128,128}

\tikzstyle{every node}=[font=\footnotesize]
\begin{axis}[
legend cell align={left},
legend style={
  fill=none,
  fill opacity=0.9,
  draw opacity=1,
  text opacity=1,
  at={(0.0,0.18)},
  anchor=north west,
  draw=none,
  column sep = 0,
  legend columns=3,
  font=\footnotesize
},
tick align=outside,
tick pos=left,
x grid style={darkgray176},
xlabel={Test-time adaptation step $(t)$},
xmin=0, xmax=500,
xtick distance=100,
ytick distance=0.1,
xtick style={color=black},
y grid style={darkgray176},
ylabel={Testing Error},
xtick align=inside,
ytick align=inside,
xticklabel style = {yshift=-1.0pt, font=\footnotesize},
yticklabel style = {xshift=4pt, , font=\footnotesize},
ymajorgrids,
xmajorgrids,
height=6.5cm, width = 8.8 cm,
ymin=0.3, ymax=0.9,
ytick style={color=black},
outer sep=0pt,
outer xsep=3pt,
every axis x label/.style={
    at={(axis description cs:0.5,-0.18)},
    align = center},
every axis y label/.style={
    rotate=90,
    at={(axis description cs:-0.1,0.5)},
    align = center},
]
\addplot [semithick, maroon, mark=x, mark size=2.5, mark options={solid}]
table {%
0 0.4886
25 0.5007
50 0.5048
75 0.5137
100 0.5182
125 0.5289
150 0.5385
175 0.5539
200 0.563
225 0.5687
250 0.5765
275 0.5902
300 0.5996
325 0.6044
350 0.6104
375 0.6211
400 0.6346
425 0.6361
450 0.6396
475 0.6464
};
\addlegendentry{$B=8$}
\addplot [semithick, darkorange, mark=+, mark size=2.5, mark options={solid}]
table {%
0 0.4854
25 0.5006
50 0.5143
75 0.5248
100 0.533
125 0.5401
150 0.5575
175 0.5654
200 0.582
225 0.596
250 0.6058
275 0.6128
300 0.6211
325 0.6308
350 0.643
375 0.6529
400 0.6661
425 0.6886
450 0.7046
475 0.7125
};
\addlegendentry{$B=16$}
\addplot [semithick, red, mark=*, mark size=2.5, mark options={solid}]
table {%
0 0.486
25 0.5087
50 0.5248
75 0.5379
100 0.5455
125 0.5571
150 0.5717
175 0.5916
200 0.6082
225 0.6265
250 0.6409
275 0.6549
300 0.6676
325 0.6838
350 0.7064
375 0.7229
400 0.7375
425 0.7545
450 0.7707
475 0.7933
};
\addlegendentry{$B=32$}
\addplot [semithick, blue, mark=square*, mark size=2.5, mark options={solid}]
table {%
0 0.4853
25 0.4971
50 0.5101
75 0.5296
100 0.5508
125 0.5809
150 0.606
175 0.636
200 0.6581
225 0.6824
250 0.7031
275 0.7204
300 0.7391
325 0.7579
350 0.7761
375 0.7906
400 0.8067
425 0.8154
450 0.8243
475 0.8335
};
\addlegendentry{$B=64$}
\addplot [semithick, purple, mark=diamond*, mark size=2.5, mark options={solid}]
table {%
0 0.4851
25 0.5111
50 0.5384
75 0.5706
100 0.5984
125 0.6174
150 0.6308
175 0.6554
200 0.671
225 0.6859
250 0.7082
275 0.7382
300 0.7529
325 0.773
350 0.788
375 0.7942
400 0.8
425 0.8042
450 0.8123
475 0.8253
};
\addlegendentry{$B=128$}
\addplot [semithick, teal, mark=pentagon*, mark size=2.5, mark options={solid}]
table {%
0 0.4865
25 0.5
50 0.5375
75 0.5726
100 0.6072
125 0.6238
150 0.6523
175 0.6819
200 0.7043
225 0.7218
250 0.7366
275 0.7418
300 0.7524
325 0.7689
350 0.7756
375 0.7877
400 0.7934
425 0.8024
450 0.8107
475 0.8259
};
\addlegendentry{$B=256$}
\end{axis}

\end{tikzpicture}
        \caption{Various choices of the testing batch size $B$.}
        \label{fig:results_batchsize_attack}
    \end{subfigure}
    \caption{Average testing error of the baseline model  (\Sec{\ref{sec:baseline_method}}) with different variations of RIP attack.}
    
\end{figure*}

RIP attack requires a dataset with labels $\mathcal{D}_a$ to perform (for checking the correctness of the model predictions). 
In the main experiments on CIFAR-10C, CIFAR-100-C, and ImageNet-C~\cite{hendrycks2019robustness}, we conveniently adopt the set of impulse-noise corrupted images (at level 5) from these datasets.
\textit{We note that this choice of corruption is arbitrary}, and one can generate a similar dataset as long as there is a distribution mismatch with the source model, and it is likely to produce incorrect predictions.
In \Fig{\ref{fig:results_corruptions_for_attack}}, we present the baseline model collapsed by 4 other choices of $\mathcal{D}_a$, formed by picking other types of corruptions: snow, JPEG compression, motion blur, and contrast adjustment. They are representatives for their group, among a total of 15 corruptions in~\cite{hendrycks2019robustness}.
We observe a similar collapse pattern on the baseline model regardless of the choice of $\mathcal{D}_a$. 

\label{sec:random_data_augmentation}
\begin{figure*}[ht!]
    \centering
    \includegraphics[width=\linewidth]{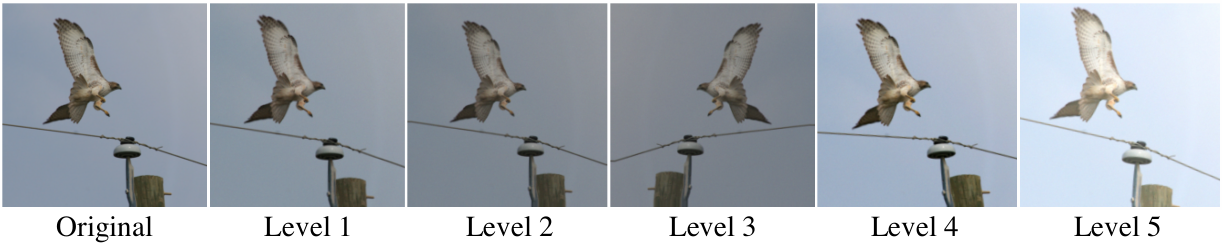}
    \caption{An example of augmented images at five different data augmentation levels.}
    \label{fig:data-aug-lvl}
\end{figure*}

\subsection{Varying Choices of Testing Batch Size}
\label{sec:different_batch_sizes}
We use a batch size of $B=64$, following previous studies~\cite{Wang_2022_CVPR, döbler2023robust, yuan2023robust} in our main experiments.
In \Fig{\ref{fig:results_batchsize_attack}}, we further investigate RIP attack on the baseline model with varying choices of $B\in \{8, 16, 32, 64, 128, 256\}$.
RIP attack consistently increases testing error across all choices of $B$, though the amount may vary with a larger batch size favors the attack.
The effect of increasing batch size seems saturated for the choices larger than $64$.

\section{Implementation Details}
\label{sec:implementation_details}

\subsection{Random Data Augmentation Operator}
Following prior work~\cite{Wang_2022_CVPR}, the image-level data augmentation \texttt{Aug}$(\cdot)$ investigated in this study is composed of a series of operators: 
\textit{random horizontal flipping, affine transformation, color jittering, additive Gaussian noise, blurring by a Gaussian kernel.}
\Sec{\ref{sec:study_level_augmentation}} introduces five level of augmentations. They are created by varying the degree of randomness and the strength of each component augmentation operator.
Specifically, we keep the probability for random horizontal flipping at $0.5$ and the kernel size of $3$, standard deviation of $0.005$ for the Gaussian blur at all levels. Other components' parameters at each level are detailed in \Tab{\ref{tab:aug_params}}.

For illustration, in \Fig{\ref{fig:data-aug-lvl}}, the effect of augmentation on an image in ImageNet~\cite{jia2009_imagenet} at each level is provided.

\begin{table*}[ht]
\resizebox{\textwidth}{!}{
\begin{tabular}{|c|c|cc|ccccc|}
\toprule
\multicolumn{1}{|c|}{\multirow{2}{*}{\textbf{Level}}} & \multicolumn{1}{c|}{\multirow{2}{*}{\textbf{\begin{tabular}[c]{@{}c@{}}Gaussian\\ Noise\end{tabular}}}} & \multicolumn{2}{c|}{\textbf{Random Affine}}                                                                             & \multicolumn{5}{c|}{\textbf{Color Jitter}}                                                                                                                                                         \\ \cline{3-9} 
\multicolumn{1}{|c|}{}                                & \multicolumn{1}{c|}{}                                                                                  & \multicolumn{1}{c}{\textbf{Degrees}} & \multicolumn{1}{c|}{\textbf{Scale}}                                              & \multicolumn{1}{c}{\textbf{Brightness}} & \multicolumn{1}{c}{\textbf{Contrast}} & \multicolumn{1}{c}{\textbf{Saturation}} & \multicolumn{1}{c}{\textbf{Hue}} & \multicolumn{1}{c|}{\textbf{Gamma}} 
\\ \midrule
1  & $ \sigma \in [0,001, 0.05] $ & $[-1, 1]$ & $[0.95, 0.97]$ & $[0.85, 0.87]$ & $[0.85, 0.90]$ & $[0.75, 0.80]$ & $[-0.005, 0.005]$ & $[0.95, 1.00]$
\\
2  & $ \sigma \in [0,001, 0.10] $ & $[-2, 2]$ & $[0.95, 1.00]$ & $[0.8, 0.9]$ & $[0.85, 0.95]$ & $[0.75, 0.85]$ & $[-0.01, 0.01]$ & $[0.95, 1.05]$
\\
3  & $ \sigma \in [0,001, 0.15] $ & $[-4, 4]$ & $[0.95, 1.05]$ & $[0.9, 1.1]$ & $[0.85, 1.05]$ & $[0.75, 1.15]$ & $[-0.02, 0.02]$ & $[0.85, 1.05]$                                    
\\
4  & $ \sigma \in [0,001, 0.15] $ & $[-8, 8]$ & $[0.95, 1.05]$ & $[0.8, 1.2]$ & $[0.85, 1.15]$ & $[0.75, 1.25]$ & $[-0.03, 0.03]$ & $[0.85, 1.05]$                                
\\
5  & $ \sigma \in [0,001, 0.25] $ & $[-15, 15]$ & $[0.9, 1.1]$ & $[0.6, 1.4]$ & $[0.7, 1.3]$ & $[0.50, 1.50]$ & $[-0.06, 0.06]$ & $[0.7, 1.3]$ 
\\ 
\bottomrule
\end{tabular}}
\caption{Parameters for the component random data augmentation operators in \texttt{Aug} at five different levels.}
\label{tab:aug_params}
\end{table*}

\subsection{The Numerical Simulation on Gaussian Mixture Model Classifier (GMMC)}
\label{sec:setup_gmmc_numerical_simulation}
For the GMMC simulation mentioned in our analysis, a toy dataset is constructed with $N=1,000$ data points, drawn from a mixture of two Gaussian distributions: $(\mu_0, \mu_1) = (-1.0, 2.0)$ and $\sigma_0=\sigma_1=1.0$.
We simulate a TTA process with $T_a = 120$ steps. The level of augmentation in the Additive White Gaussian Noise (AWGN) opearator is set to $\sigma=0.2$.  
Visit \cite{hoang2024petta} for further discussions on GMMC.

\subsection{Continual Test-time Adaptation Methods}
\noindent \textbf{Source Model and Dataset. } 
RobustBench~\cite{croce2021robustbench} and \texttt{torchvision}~\cite{torchvision2016} provide the initial models ($f_0$) trained on the source distributions. From \texttt{RobustBench}, the model with checkpoint \texttt{Standard} and \texttt{Hendrycks2020AugMix\_ResNeXt}~\cite{hendrycks2020augmix} are adopted for CIFAR10-C and CIFAR-100-C experiments, respectively. The ResNet50~\cite{He2015} model pre-trained on ImageNet V2 (specifically, checkpoint \texttt{ResNet50\_Weights.IMAGENET1K\_V2} of \texttt{torchvision}) is used for ImageNet-C experiments.

\noindent \textbf{Updated Parameters. }
Following prior studies~\cite{wang2021tent, yuan2023robust, döbler2023robust, Wang_2022_CVPR}, $\theta_t$ - the linear parameters of batch norm layers~\cite{pmlr-v37-ioffe15} are updated at each adaptation step $t$. 

\noindent \textbf{Optimizer. } Adam~\cite{KingmaB14} optimizer with learning rate equal $1e^{-3}$, and $\beta = (0.9, 0.999)$ is selected as a universal choice for all experiments. 

\subsection{RIP Attack Trials}
As mentioned in the experimental setup, the average performance across $10$ RIP attack trials is reported. 
For each trial, all settings are kept the same, except for the victim class $y_a$.
This class is uniformly sampled without replacement from all possible classes. For reproducibility, we list the index of the victim classes $(y_a)$ used in our experiments:
\begin{itemize}
    \item CIFAR-10-C: $0, 1, 2, 3, 4, 5, 6 ,7, 8 , 9, 10$ \textit{(all classes)}.
    \item CIFAR-100-C: $3, 8, 29, 48, 56, 67, 71, 88, 91, 96$.
    \item ImageNet-C: $91, 323, 392, 583, 630, 637, 643, 707, 864, 952$.
\end{itemize}

\subsection{Computing Resources}
\label{sec:computing_resources}

Experiments are conducted on a computer cluster with an Intel(R) Core(TM) $3.30$GHz Intel Core i9-9820X CPU, 128 GB RAM, and $4\times$NVIDIA Quadro RTX 5000 GPUs.

\end{document}